\documentclass[10pt,twocolumn,letterpaper]{article}

\usepackage{iccv}
\usepackage{times}
\usepackage{epsfig}
\usepackage{graphicx}
\usepackage{amsmath}
\usepackage{amssymb}
\usepackage{colortbl}
\usepackage{color}
\usepackage{subfigure}          % for sub-figure
\usepackage{rotating}
\usepackage{multirow}           % for complex table
\usepackage{bm}
\usepackage{lineno}
\usepackage{tabularx}
\usepackage{algorithm} %require ensure
\usepackage{algorithmic} %require ensure
\usepackage{amsthm}
\usepackage{url}

\definecolor{mygray}{gray}{.9}
\def\txtred#1{\textcolor{red}{\textbf{#1}}}
\def\txtblu#1{\textcolor{blue}{\textit{#1}}}

\iccvfinalcopy % *** Uncomment this line for the final submission

\def\iccvPaperID{****} % *** Enter the ICCV Paper ID here

\ificcvfinal\fi
\begin{document}

%%%%%%%%% TITLE
\title{Temporal Dynamic Appearance Modeling for Online Multi-Person Tracking}
%\title{Exploiting Temporal Dynamics on Nolinear Appearance Manifold for Online Multiple People Tracking}

\author{Min Yang \quad Yunde Jia\\
Beijing Laboratory of Intelligent Information Technology\\
School of Computer Science, Beijing Institute of Technology, Beijing 100081, P.R. China\\
{\tt\small \{yangminbit,jiayunde\}@bit.edu.cn}
%\and
}

\maketitle
%\thispagestyle{empty}

%%%%%%%%% ABSTRACT
\begin{abstract}
Robust online multi-person tracking requires the correct associations of online detection responses with existing trajectories. We address this problem by developing a novel appearance modeling approach to provide accurate appearance affinities to guide data association. In contrast to most existing algorithms that only consider the spatial structure of human appearances, we exploit the temporal dynamic characteristics within temporal appearance sequences to discriminate different persons. The temporal dynamic makes a sufficient complement to the spatial structure of varying appearances in the feature space, which significantly improves the affinity measurement between trajectories and detections. We propose a feature selection algorithm to describe the appearance variations with mid-level semantic features, and demonstrate its usefulness in terms of temporal dynamic appearance modeling. Moreover, the appearance model is learned incrementally by alternatively evaluating newly-observed appearances and adjusting the model parameters to be suitable for online tracking. Reliable tracking of multiple persons in complex scenes is achieved by incorporating the learned model into an online tracking-by-detection framework. Our experiments on the challenging benchmark MOTChallenge 2015 \cite{leal2015motchallenge} demonstrate that our method outperforms the state-of-the-art multi-person tracking algorithms.
\end{abstract}

%%%%%%%%% BODY TEXT
\section{Introduction}

Online multi-person tracking aims to estimate the trajectories of multiple persons based on the observations up to the current frame, and output the results on the fly (\ie, without temporal delay).
Driven by the significant progresses in human detection \cite{dalal2005histograms,felzenszwalb2010object,dollar2014fast}, \emph{tracking-by-detection} technique \cite{wu2007detection,xing2009multi,breitenstein2011online,leal2014learning,geiger20143d,yoon2015bayesian} has become increasingly popular. Consequently, online multi-person tracking is formulated as a data association problem where detection responses need to be reliably linked with existing trajectories frame-by-frame.
Although much progress has been reported, it is still a very challenging problem to track multiple persons in complex scenes, due to the presence of background clutters, occlusions, and frequent interactions among targets.

Finding correct associations heavily relies on the affinity measurement between a trajectory and a detection. Appearance plays an important role in the recognition of different individuals, and thus is a critical cue for affinity measurement. Much work of online multi-person tracking has been devoted to the design of good appearance models \cite{breitenstein2011online,shu2012part,kim2013online,BaeY2014robust}.
These methods aim to learn the discriminativeness among targets by solving classification problems, and show impressive tracking performance in unconstrained environments. Our work is inspired by this progress, and focuses on learning an effective appearance model for robust online multi-person tracking.

%Instead of using simple features (\eg, color histograms \cite{wu2007detection,xing2009multi,yoon2015bayesian}) to represent the tracked persons, these methods aim to learn the discriminativeness among targets by solving classification problems. However, due to the limitation that all targets come from the same class (\ie, human), the appearances of different targets could draw from similar underlying data distributions. The appearance models learned by conventional classification algorithms are prone to loose the discriminative ability, and cause ambiguous associations.

\begin{figure*}[ht]
    \centering
    \includegraphics[width=0.98\textwidth]{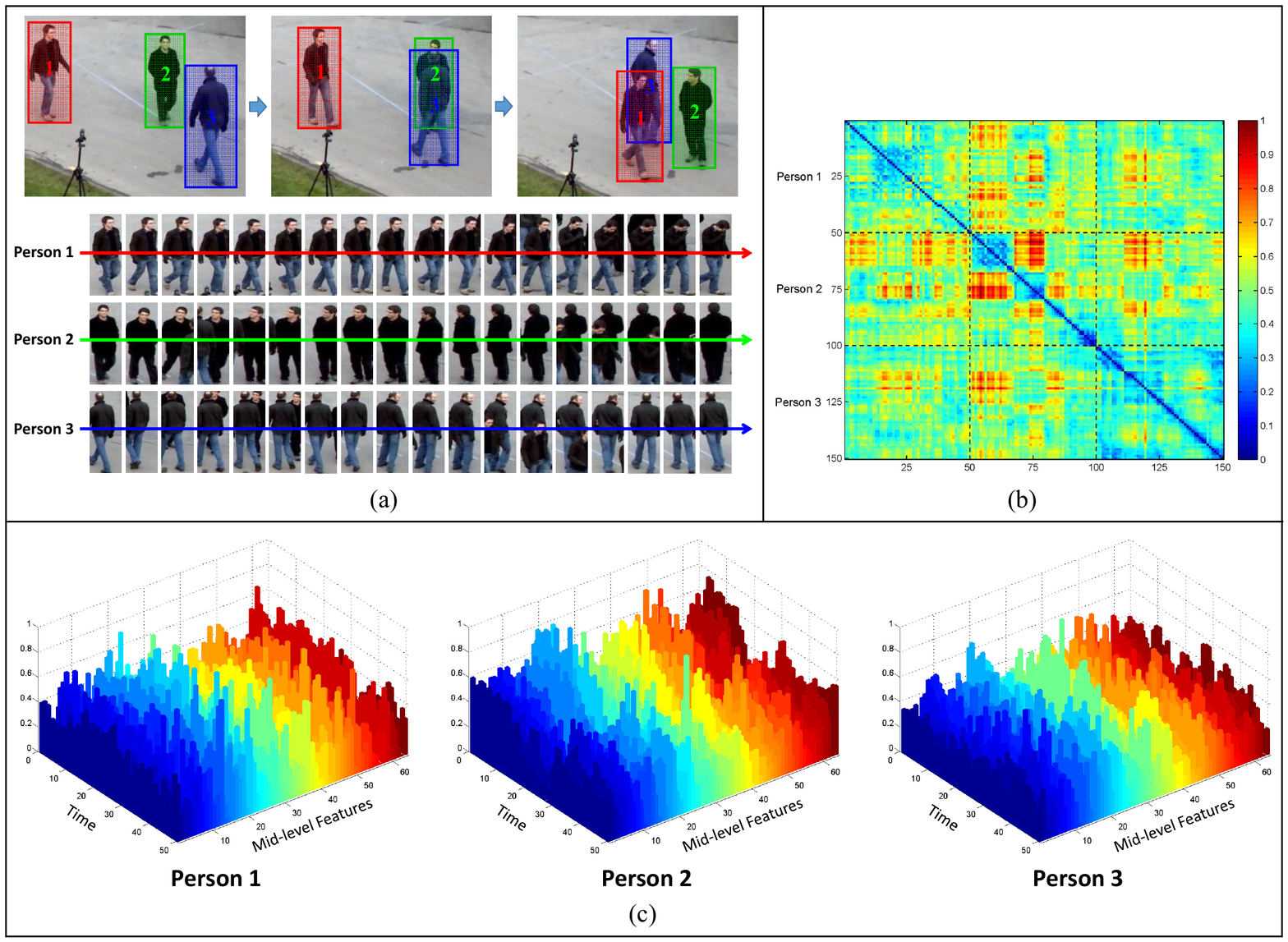}
    \vspace{-0pt}
    \caption{Illustration of the basic idea in our work. We show the groundtruth appearances of three closely interacting people (marked as red, green, and blue colors, respectively) in (a), and visualize the distances between these appearances (50 successive frames for each person) in the mid-level feature space (see Sec.~\ref{pretrain}) in (b). While there is some block structure, significant confusion between individuals can be found when the dynamic appearances of each person is considered as an unordered collection. In contrast, the temporal dynamic characteristics within each appearance sequence, illustrated by the temporal evolution of mid-level feature vectors in (c), are helpful to discriminate different individuals. }
    \vspace{-0pt}
    \label{motivation}
\end{figure*}

We explore the temporal dynamic to enhance the appearance model. The basic idea is that the appearances of one person during tracking can be considered as a temporal sequence with dependencies across times, rather than a discrete set that only involves the spatial structure in the feature space. As shown in Fig.~\ref{motivation}, the temporal dynamic characteristics within appearance sequences can be exploited to disambiguate different individuals, even when they have similar appearances at some times. Incorporating the temporal dynamic into appearance modeling significantly improves the accuracy of affinity measurement, and can exhibit its usefulness in facilitating multi-person tracking.

In this paper, we propose an algorithm for learning temporal dynamic appearance model (TDAM) which is designed to automatically find the person-specific temporal dynamic to improve online multi-person tracking performance. Specifically, we employ the Hidden Markov Model (HMM) to capture both the temporal dynamic and the spatial structure when the appearances of persons change over time. To benefit the characterizing of the temporal dynamic, the redundant low-level features of human appearance are mapped into a mid-level semantic feature space by using a feature selection approach. An online Expectation-Maximization (EM) algorithm is utilized to incrementally update the model parameters, which makes the TDAM suitable for online tracking. Moreover, we do careful design on model initialization to ensure the effectiveness of incremental learning. The TDAM is incorporated into an online tracking-by-detection framework to obtain robust multi-person tracking results in complex scenes. By considering the spatial similarity as well as the temporal dependencies within appearance sequences, the appearance affinities between trajectories and detections provided by the TDAM are more accurate and reliable. Evaluation on a new challenging benchmark (MOTChallenge 2015 \cite{leal2015motchallenge}) demonstrates that our method outperforms the state-of-the-art trackers.

In summary, this paper makes the following contributions:
\begin{itemize}\setlength{\itemsep}{-0pt}

\item We propose a temporal dynamic appearance modeling approach to effectively capture both the temporal dynamic and the spatial structure of dynamic human appearances. To learn the model online, we present an incremental learning algorithm which alternatively evaluates newly-observed appearances and adjusts the model parameters.

\item We describe a feature selection algorithm to map the redundant low-level features of human appearance into a low-dimensional mid-level feature space. We show that the mid-level features provide richer semantic interpretation of appearance changes and therefore benefit temporal dynamic appearance modeling.

\item We propose a novel online multi-person tracking method which explicitly exploits the temporal dynamic characteristics of varying appearances to ensure the correctness of data association. Compared to conventional methods, our method is more robust against frequent and close interactions between persons.
\end{itemize}

The rest of the paper is organized as follows. Section 2 reviews the related work. Section 3 presents the details of the TDAM algorithm, and Section 4 describes the implementation of our online multi-person tracking method. We report and discuss the experimental results in Section 5, and conclude the paper in Section 6.

\section{Related Work}

%The major issue of multi-person tracking-by-detection methods is how to correctly associate detection responses which belong to the same person.

Appearance modeling has gained increasing attention in the literature of multi-person tracking. Many algorithms solve the problem in a large temporal window, and perform global optimization (\eg, linear programming \cite{ben2011tracking}, condition random field \cite{yang2012multi,choi2015near}, or continuous energy minimization \cite{Milan:2014:CEM}) of multiple trajectories, where appearance information is formulated as constraints in the cost function.
Kuo \etal \cite{kuo2010multi} use online-trained classifiers to estimate the appearance affinity between two short term trajectories (tracklets), and obtain multi-person tracking using a hierarchical association framework. This work is extended in \cite{kuo2011does,yang2012multi} to utilize both target-specific and global appearance models for tracklets association. Brendel \etal \cite{brendel2011multiobject} and Wang \etal \cite{wang2014tracklet} employ distance metric learning to find appropriate match between the appearances of detections or tracklets. Due to the significant temporal delay and time-consuming iterative process, it is difficult to apply these methods to time-critical applications.

Our work focuses on building an effective appearance model for online multi-person tracking, which only considers observations up to the current frame and outputs trajectories without temporal delay. The conventional approach of appearance modeling is using descriptors, such as color histograms \cite{wu2007detection,xing2009multi,poiesi2013multi,yoon2015bayesian}, to represent the targets, and compute the similarities between descriptors to indicate appearance affinities. Alternatively, Yang \etal \cite{yang2009detection} use multi-cue integration to fuse color, shape and texture features to build a more sophisticated appearance model. Breitenstein \etal \cite{breitenstein2011online} and Shu \etal \cite{shu2012part} integrate person-specific classifiers to obtain discriminative appearance models that are able to distinguish the tracked person from the background and other targets. Bae and Yoon \cite{bae2014robust} use online-trained classifiers to evaluate the observation likelihood in terms of appearance information, and perform multi-person tracking within a Bayesian framework. Kim \etal \cite{kim2013online} use structured SVM to predict the appearance similarity between all pairs of targets in two consecutive frames. Bae and Yoon \cite{BaeY2014robust} employ incremental linear discriminant analysis to find a projection that gathers the appearances from the same target and simultaneously separates the appearances from different targets.

The methods mentioned above only consider the spatial structure of dynamic appearances in the feature space, and might produce ambiguous results when the appearances of different persons are similar at some static times. Since the targets of interest in multi-person tracking come from the same category (\ie, human), the appearance sequences of different people may exhibit similar spatial structure. Conventional appearance models, therefore, usually degrade especially when the targets are interacting within close proximity. In contrast, the TDAM proposed in this paper exploits the temporal dynamic characteristics of appearance variations to enhance the appearance model. We show that the temporal dynamic makes a sufficient complement to the spatial structure of varying appearances in terms of distinguishing individuals. The appearance affinities computed by the TDAM are more reliable for data association between trajectories and detections, and thus facilitates multi-person tracking.

The concept of temporal dynamic is not new in the community of computer vision. It has been introduced into action/event recognition \cite{li2013recognizing,bhattacharya2014recognition}, dynamic appearance prediction \cite{lim2005caratheodory,lim2006dynamic}, and video-based face recognition \cite{liu2003video,lee2003video}. To our best knowledge, ours is the first work to model temporal dynamic characteristics of dynamic appearances for multi-person tracking. We demonstrate its significant importance and practical usefulness in online multi-person tracking by solving crucial issues including feature selection, appearance matching, and model parameter estimation.

\begin{figure}[t]
    \centering
    \includegraphics[width=0.45\textwidth]{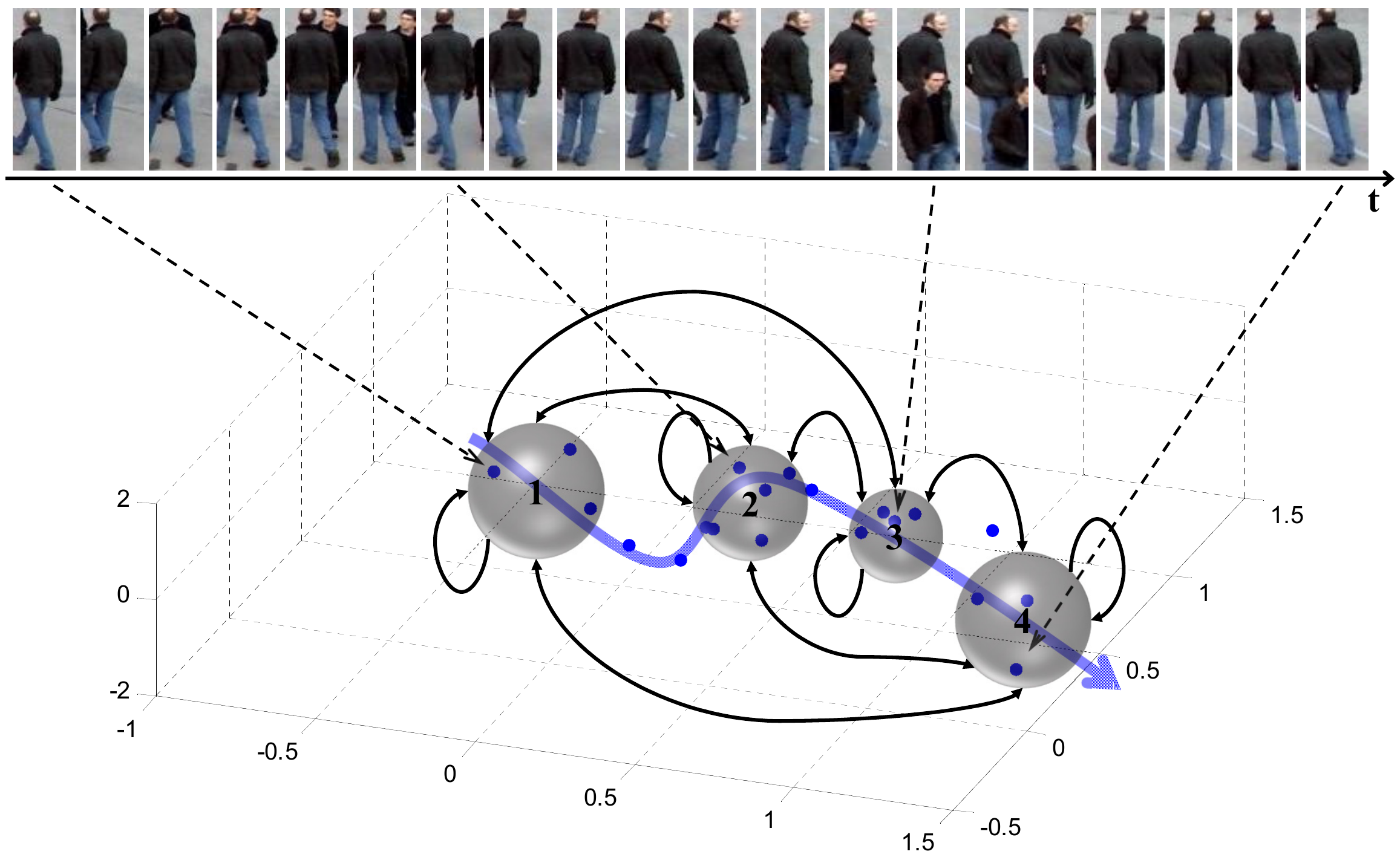}
    \caption{Illustration of using the TDAM to model an appearance sequence. Hidden states are indicated by gray spheres, and the state transitions are represented by directed edges. The temporal dynamic of the appearance sequence is illustrated as a blue arrow. }
    \vspace{-0pt}
    \label{TDAM-Example}
\end{figure}

\section{Temporal Dynamic Appearance Modeling}

\subsection{Background}
\label{background}

Given an appearance sequence of one specific person, we build a temporal dynamic appearance model (TDAM) by using the Hidden Markov Model (HMM) \cite{rabiner1989tutorial} to characterize the temporal dynamic as well as the spatial structure of appearances in the feature space. The assumption of using the HMM for appearance modeling is that the appearance sequence is associated with an unobservable Markov chain with a finite number of states. The appearance at each time only depends on the observation density of the associated hidden state. In this way, the temporal dynamic of appearance changes is indicated by the underlying Markov process, and the spatial structure is captured by multiple observation densities. An example of the TDAM is illustrated in Fig.~\ref{TDAM-Example}.

Formally, we represent the appearance and the associated hidden state at time $t$ as $\mathbf{o}_t \in \mathbb{R}^d$ and $s_t \in \{1,\ldots,N\}$, respectively, where $d$ is the dimensionality of the feature space and $N$ is the number of states. A TDAM can be parameterized as the HMM parameters $\bm{\theta} = (\bm{\pi}, \mathbf{A}, \mathbf{F})$, where
\begin{itemize}\setlength{\itemsep}{-0pt}
\item $\bm{\pi} = \{\pi_i\}$ is the initial state distribution,
    \begin{equation}\nonumber
    \pi_i = P\left(s_1 = i\right), \quad i = 1,\ldots,N.
    \end{equation}
\item $\mathbf{A} = \{a_{ij}\}$ is the state transition matrix of the Markov process,
    \begin{equation}\nonumber
    a_{ij} = P\left(s_{t+1} = j | s_t = i\right), \quad i,j = 1,\ldots,N,
    \end{equation}
    where the state transition probabilities satisfy the constraint $\sum_{j=1}^{N}{a_{ij}} = 1, i = 1,\ldots,N$.
\item $\mathbf{F} = \{f_i(\cdot)\}$ is the observation densities associated with hidden states. A common choice is Gaussian mixture models (GMMs),
    \begin{align}\label{eq:1-0}
    f_i\left(\mathbf{o}_t\right) &= P\left(\mathbf{o}_t | s_t = i\right) \nonumber \\
     &= \sum_{k=1}^{M}{\omega_{ik} \cdot \mathcal{N}\left( \mathbf{o}_t;\bm{\mu}_{ik}, \Sigma_{ik} \right)},
    \end{align}
    where each observation density $f_i(\cdot)$ is represented by a mixture of $M$ Gaussian distributions, and $\omega_{ik}$, $\bm{\mu}_{ik}$, and $\Sigma_{ik}$ are the mixture weight, the mean vector, and the covariance matrix of $k$-th component, respectively.
\end{itemize}

There are two critical problems in appearance modeling: model learning for the estimation of model parameters, and appearance matching for the evaluation of new appearances. Usually, with a set of training sequences, the model parameters of an HMM are estimated by using maximum likelihood estimation \cite{juang1986maximum}. Then the likelihood that an test sequence is generated by the model can be evaluated via a forward procedure \cite{rabiner1989tutorial}. However, applying this common strategy to the TDAM is infeasible, because we build the trajectory of each person sequentially, the whole appearance sequence is not available at once. In this work, we divide the appearance sequence into fixed-length subsequences using temporal sliding windows, and turn to model the temporal dynamic of appearance changes for each person within a short time. Appearance modeling is then performed in an online manner. That is, at each time, a TDAM evaluates new appearances with its current parameters, and then incrementally updates the parameters to adapt itself to newly obtained training data.

Although such a strategy appears desirable, it raises several issues. Firstly, the low-level features (\eg, color and shape) that represent human appearance are often high-dimensional, while the number of training data, restricted by tracking application, is relatively small. It makes the estimation of TDAM parameters extremely difficult. In addition, due to the large number of TDAM parameters, incremental update of the model might be sensitive to initialization.
In order to make the online learning of the TDAM effective, we present a feature selection approach to map high-dimensional low-level features into low-dimensional mid-level features. We show that the temporal dynamic characteristics can be well captured by representing an appearance sequence as a vector time-series, whose elements correspond to temporal evolution of different semantic features. In terms of model initialization, we learn a general TDAM off-line to roughly character the spatial and temporal properties of human appearances. Person-specific properties are retained during online learning to evolve the general TDAM to a person-specific TDAM. In the following subsections, we present the details of the design.

\subsection{Feature Selection}
\label{pretrain}

Human appearance is usually described by using low-level features such as color, shape and texture. However, it is extremely difficult to directly model the temporal dynamic characteristics of low-level features, as they are redundant and sensitive to noise. In contrast, mid-level features (\eg, bag of words \cite{li2013recognizing} and detector confidence \cite{bhattacharya2014recognition}) provide a richer semantic interpretation and facilitate the exploitation of temporal dynamics. We thus propose to extract meaningful mid-level features from original low-level features for the online learning of the TDAM. Specifically, we learn a set of detectors in an off-line process, each of which can provide the probability that one person exhibits a specific category of appearance. The resulting detectors are similar to \emph{attribute} or \emph{concept} detectors which are popular in the community of action/event recognition and have shown impressive results.

To learn such detectors, a large amount of static human appearances from the INRIA dataset~\cite{dalal2005histograms} are exploited as training data. For each human appearance, we normalize the image window into $128 \times 48$ pixels and decompose it into cells of $8 \times 8$ pixels. An augmented HOG feature vector \cite{felzenszwalb2010object} with dimensionality $31$ is then extracted from each cell. The whole feature vector that represents one human appearance thus has dimensionality $16 \times 6 \times 31 = 2976$. We choose the HOG feature as the low-level representation of human appearances as it offers an effective way to describe the shape information, which is more robust to noise than other features such as color and texture.

\textbf{Clustering:} Given the training data, our approach first aggregates similar human appearances into appearance clusters by using recursive normalized cuts \cite{shi2000normalized}. Then one candidate detector is trained for each cluster to distinguish the appearances associated with this cluster from others. To avoid hard negative mining for each trained detector, we use the LDA technique of \cite{hariharan2012discriminative} to learn a soft but universal model of `negative' appearance. The parameter vector $w$ of a candidate detector (expressed as a linear classifier) is simply learned by
\begin{equation}\label{eq:0-1}
    w = \Sigma^{-1}\left(\bar{x} - \mu_{0} \right),
\end{equation}
where $\bar{x}$ is the mean of the HOG features of the appearances in the cluster, $\mu_{0}$ and $\Sigma$ are, respectively, the mean and the covariance matrix estimated from all static human appearances in the dataset.

Note that both $\mu_{0}$ and $\Sigma$ are calculated only \emph{once} during the training process, which significantly enhances the training efficiency. Moreover, $\mu_{0}$ and $\Sigma$ can be used to whiten the HOG features to highlight the discriminative characteristics against the universal appearance \cite{hariharan2012discriminative}. We integrate this advantage into the clustering stage, and use the cosine of the angle between two whitened HOG features to indicate the affinity for normalized cuts. This strategy makes our clustering process conservative. That is, it only assigns the human appearances with exactly similar shape characteristics into the same cluster, which ensures the discriminative ability of the trained detector.

\textbf{Selection:} Given the candidate detectors, we aim to find the most `meaningful' detectors for temporal dynamic appearance modeling. The notion of a `meaningful' detector is that it should identify a specific category of human appearance from the trajectories of different persons, but not have response to other appearances. For example, a meaningful detector would prefer to identify a specific pose of one person than to just discriminate the person from the background.

To select meaningful detectors, we design an effective measure to capture this notion. We randomly choose $100$ groundtruth trajectories (longer than $50$ fames) from the training videos of the MOTChallenge 2015 benchmark for validation. Each candidate detector is evaluated on the trajectories to analyze the occurrences of the appearance cluster indicated by the detector. Denote a trajectory as $X = \{ x_{1},\ldots,x_{n} \}$, where $x_{i}$, $i \in \Omega = \{ 1,\ldots,n \}$, is the HOG feature of the $i$-th appearance in the trajectory, and $n$ is the length of the trajectory. Let $\phi_{i}$ represent the event that the $i$-th appearance in the trajectory $X$ is an occurrence of the appearance cluster indicated by the detector $w$, we can estimate the probability of the event $\phi_{i}$ given the trajectory $X$ and the detector $w$ as
\begin{equation}\label{eq:0-2}
    P(\phi_{i}|X,w) = \frac{s(x_{i}, w)}{\sum_{i=1}^{n}s(x_{i}, w)},
\end{equation}
where $s(x_{i}, w)$ is the detection score (truncated to be larger than zero) by evaluating the detector $w$ on the appearance $x_{i}$. Then the entropy $H(\Omega|X,w)$ is computed to indicate the certainty of occurrences,
\begin{equation}\label{eq:0-3}
    H(\Omega|X,w) = - \sum_{i \in \Omega}P(\phi_{i}|X,w) \log P(\phi_{i}|X,w).
\end{equation}
A lower entropy indicates a higher certainty of occurrences and thus is preferred. The average entropy over all validation trajectories is taken as an overall measure of the `meaningfulness' of a candidate detector. The top scoring detectors based on this measure are then retained.

\begin{figure*}[t]
    \centering
    \subfigure[]{
    \includegraphics[width=0.23\textwidth]{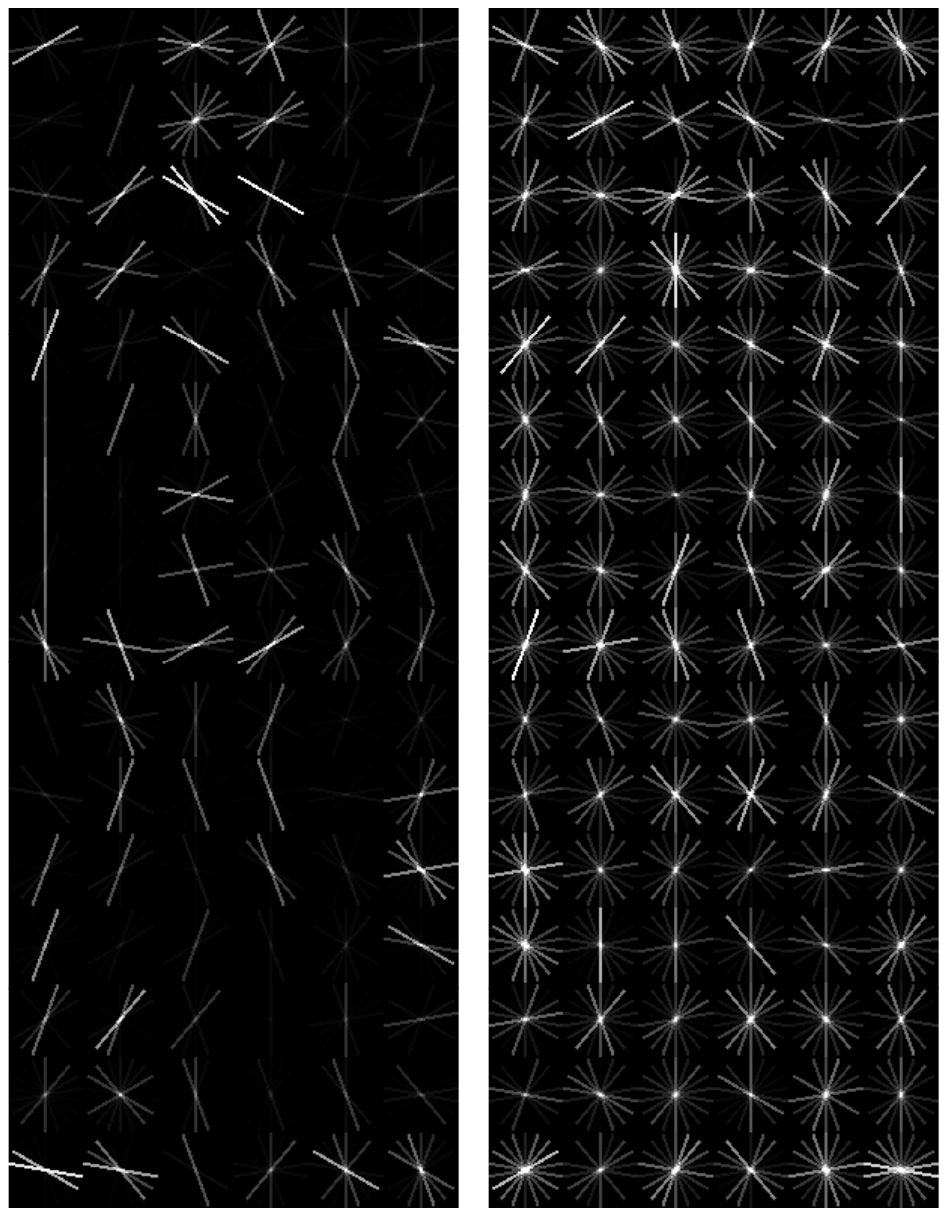}}
    \subfigure[]{
    \includegraphics[width=0.23\textwidth]{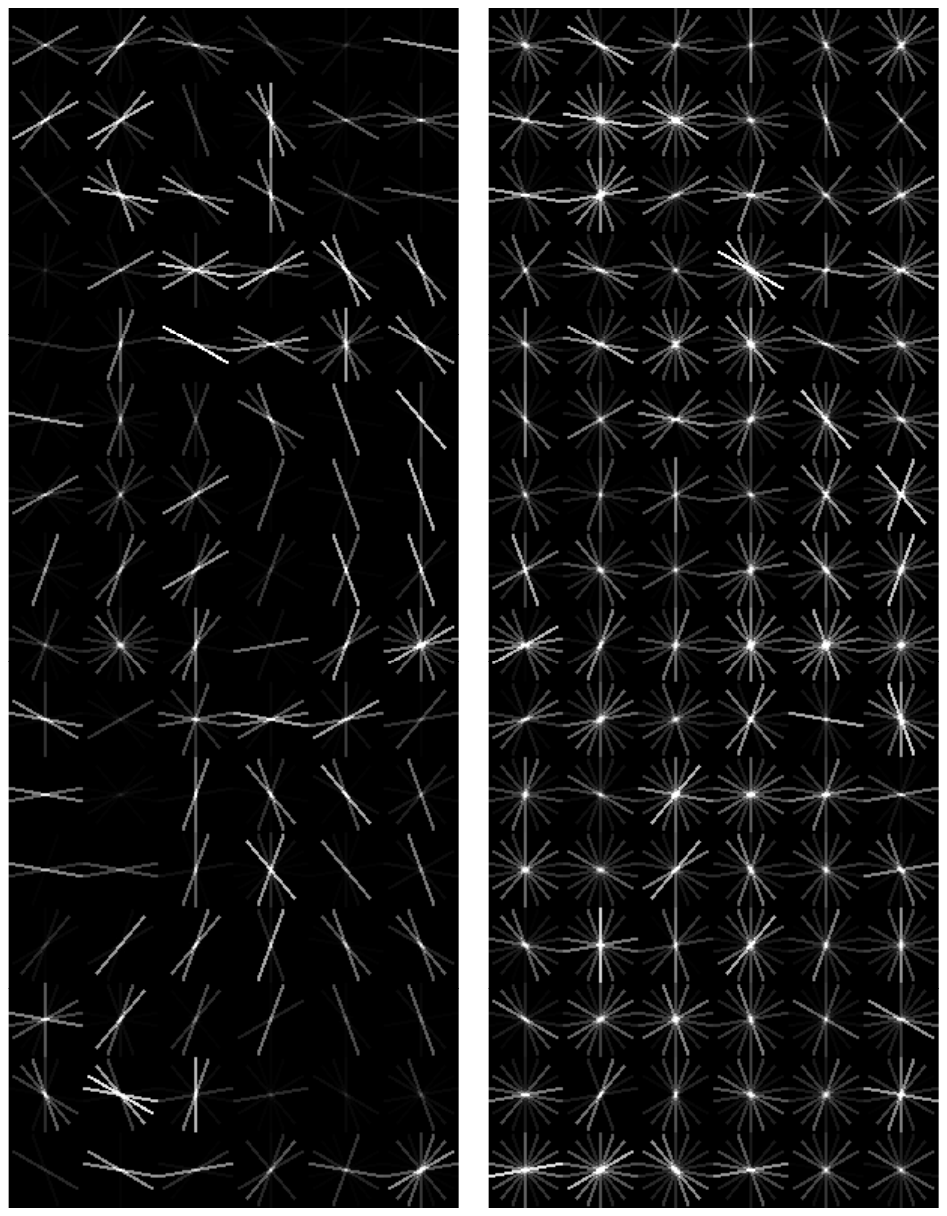}}
    \subfigure[]{
    \includegraphics[width=0.23\textwidth]{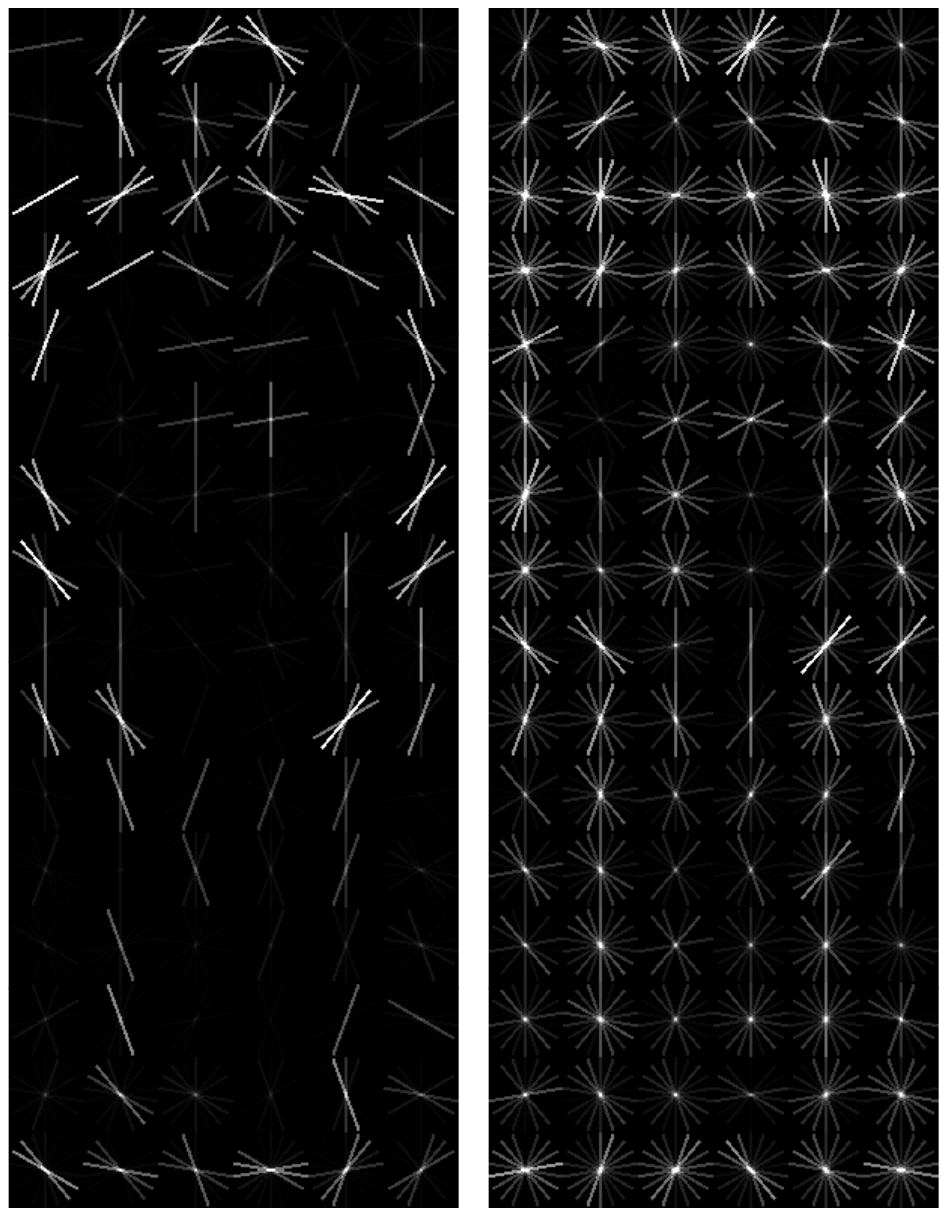}}
    \subfigure[]{
    \includegraphics[width=0.23\textwidth]{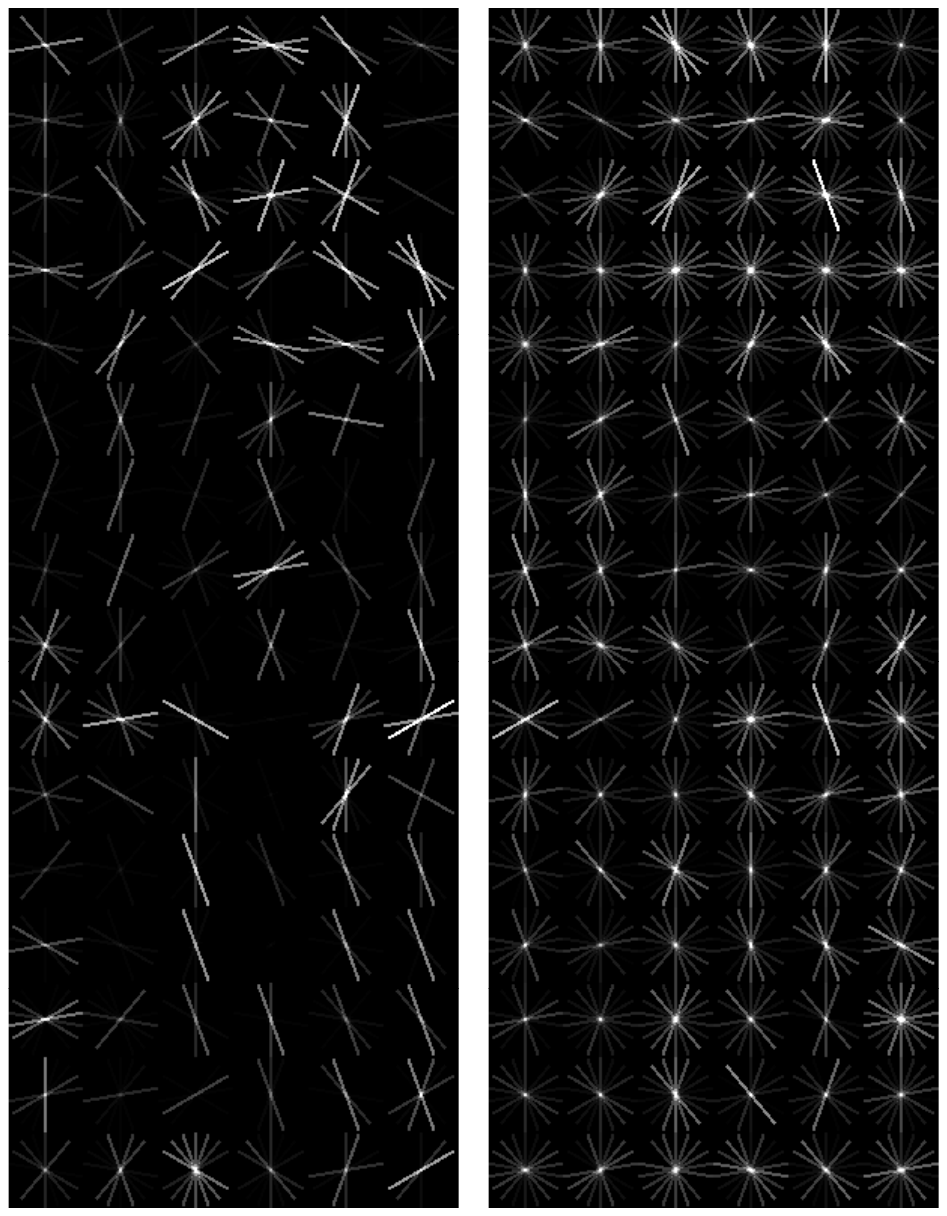}}
    \caption{Visualization of the first four detectors selected by our feature selection approach. The visualizations show the positive and negative components separately, with the positive components on the left and negative on the right.}
    \vspace{-0pt}
    \label{MidFeature-Example}
\end{figure*}

The final step is to remove redundant detectors. In fact, there is no guarantee that the selection stage described above will not return similar detectors multiple times. The redundancy between a pair of detectors can be measured by the cosine of the angle between their parameter vectors. At last $d$ detectors are selected sequentially by increasing the average entropy scores, skipping detectors that have cosine similarity larger than a threshold (set to $0.8$ in our implementation) with any of the detectors already selected. Fig.~\ref{MidFeature-Example} visualizes the first four detectors selected by our feature selection approach. As is clear from the figure, the selected detectors learn to stress general and critical person contours, and together provide well semantic interpretation of temporal appearance sequences.

With these off-line trained detectors, the HOG feature vector of one human appearance is transformed into a $d$-dimensional feature space, by taking the detection scores as mid-level semantic features. Note that this mapping is extremely efficient, because the learned detectors can be regarded as a set of linear filters.

\subsection{Appearance Matching}
\label{matching}

At each time $t$, we denote the TDAM of a specific person as $\bm{\theta}_t = \left(\bm{\pi}^{(t)}, \mathbf{A}^{(t)}, \mathbf{F}^{(t)}\right)$. Note that we drop the notion for different people since appearance modeling is independent for each person. To evaluate a new appearance $\mathbf{o}_{t+1} \in \mathbb{R}^d$ at time $(t+1)$, which has been projected into the mid-level feature space as described in Sec.~\ref{pretrain}, we compute the likelihood $P\left( \mathbf{o}_{t+1} | \bm{\theta}_t \right)$ that $\mathbf{o}_{t+1}$ is generated by the model $\bm{\theta}_t$.

\begin{figure*}[t]
    \centering
    \includegraphics[width=0.98\textwidth]{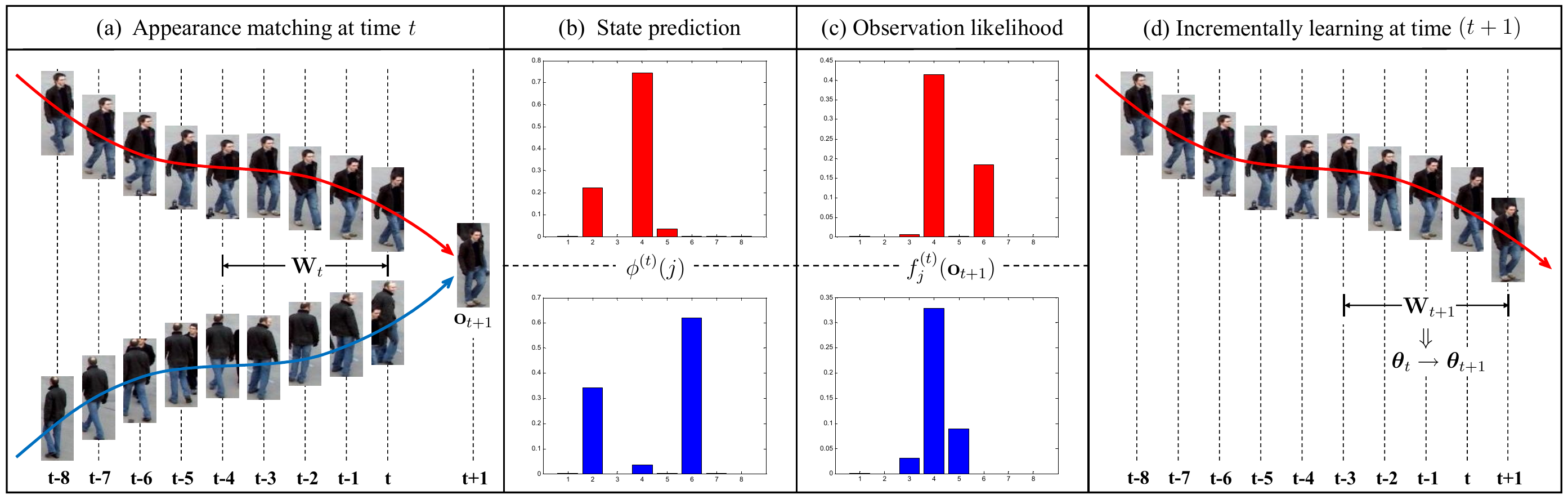}
    \caption{Illustration of the online-learning process of the TDAM. (a) A new appearance $\mathbf{o}_{t+1}$ is matched with two similar-looking persons (marked as red and blue colors, respectively) at time $t$. Two person-specific TDAMs with $N=8$ hidden states are learned online to model the varying appearances. (b) and (c), respectively, shows the state prediction probabilities and observation likelihoods provided by two TDAMs with corresponding colors. Even though the new appearance has high observation likelihoods with both of the two appearance sequences, the state prediction probabilities provide robust guide information to achieve correct matching. (d) The TDAM parameters are incrementally updated with the newly-observed appearance. }
    \vspace{-0pt}
    \label{Matching-Example}
\end{figure*}

As the TDAM $\bm{\theta}_t$ models the temporal dependencies of appearances across times, we exploit the recent appearances within a temporal window of the trajectory to analyze the temporal characteristics (see an example in Fig.~\ref{Matching-Example}(a)). Formally, we have
\begin{equation}\label{eq:1}
    P\left( \mathbf{o}_{t+1} | \bm{\theta}_t \right) = P( \mathbf{o}_{t+1} | \mathbf{W}_t, \bm{\theta}_t ),
\end{equation}
where $\mathbf{W}_t = \{ \mathbf{o}^{(t)}_{t_1},\ldots,\mathbf{o}^{(t)}_{t_L} \}$ is a subsequence containing the most recent $L$ observed appearances from the trajectory before time $t$. Due to occlusions or missed detections, the tracked person might be invisible (\ie, has no associated detections) for some times. To alleviate the influence of noise, we only exploit visible appearances in $\mathbf{W}_{t}$. Eq.~(\ref{eq:1}) implies that the new appearance $\mathbf{o}_{t+1}$ is evaluated by the TDAM $\bm{\theta}_t$ on condition of being the successor of the subsequence $\mathbf{W}_{t}$.

Using the fact that the probability of observing $\mathbf{o}_{t+1}$ only depends on the hidden state $s_{t+1}$ at time $(t+1)$, we can compute the likelihood $P( \mathbf{o}_{t+1} | \mathbf{W}_{t}, \bm{\theta}_t )$ as
\begin{eqnarray}\label{eq:2}
\begin{split}
    P( \mathbf{o}_{t+1} | \mathbf{W}_{t}, \bm{\theta}_t )
     &= \sum_{j=1}^{N}{P\left(\mathbf{o}_{t+1},s_{t+1} = j|\mathbf{W}_{t},\bm{\theta}_t\right)} \\
     &= \sum_{j=1}^{N}{\phi^{(t)}(j) \cdot f^{(t)}_{j}(\mathbf{o}_{t+1})},
\end{split}
\end{eqnarray}
where $\phi^{(t)}(j) = P\left(s_{t+1} = j|\mathbf{W}_{t},\bm{\theta}_t\right)$ is the state prediction probability, and $f^{(t)}_{j}(\mathbf{o}_{t+1}) = P\left(\mathbf{o}_{t+1}|s_{t+1} = j,\bm{\theta}_t\right)$ is the observation likelihood computed by Eq.~(\ref{eq:1-0}). The state prediction probability can be obtained by applying a forward procedure \cite{rabiner1989tutorial} on the subsequence $\mathbf{W}_{t}$ using the current model $\bm{\theta}_t$. Note that the temporal dependencies between the subsequence $\mathbf{W}_{t}$ and the new appearance $\mathbf{o}_{t+1}$ are indicated by $\phi^{(t)}(j)$, and $f^{(t)}_{j}(\mathbf{o}_{t+1})$ only depend on the observation densities in the model $\bm{\theta}_t$.

We show that using $P( \mathbf{o}_{t+1} | \mathbf{W}_{t}, \bm{\theta}_t )$ in Eq.~(\ref{eq:2}) to match appearances introduces several advantages over existing methods. First, multiple observation densities (\ie, $f^{(t)}_{j}(\mathbf{o}_{t+1})$) are exploited to model the spatial distribution of dynamic appearances. It allows the TDAM to effectively capture the appearance changes of the tracked person. Second, the temporal dynamic of appearances (\ie, $\phi^{(t)}(j)$) is explicitly involved. The TDAM learns the temporal dependencies between adjacent appearances from previous observations, and provides useful information to guide the evaluation of new appearances. In addition, the state prediction probabilities $\phi^{(t)}(j)$ are independent of new appearances, and thus only need to be computed once when many appearances are evaluated by one model. This property ensures the efficiency of using the TDAM for tracking multiple people.

To further demonstrate the importance of temporal dynamic, an example of appearance matching between a new appearance and two similar-looking people is shown in Fig.~\ref{Matching-Example}. By only considering the spatial distribution, the new appearance has high probabilities to be observed by both of the two sequences (see Fig.~\ref{Matching-Example}(c)). Most existing appearance models provide ambiguous results in this situation. In contrast, the TDAM integrates temporal dynamic to achieve more robust matching. As shown in Fig.~\ref{Matching-Example}(b), even though the new appearance is likely to be observed by the $4$-th state of the `blue' sequence, the state prediction probabilities show that the appearance tend to be originated from the $2$-nd or the $6$-th state next time. The state prediction probabilities produced by the correct sequence (\ie, the `red' sequence), by contrast, are consistent with the observation likelihoods.

\subsection{Incremental Learning}
\label{incremental}

With the appearance matching approach, a person-specific TDAM can be used to evaluate the appearance affinities between the person trajectory and online-provided detections. Based on the appearance affinities, data association is performed to link detections with appropriate trajectories, as described in Sec.~\ref{DA}. Once the trajectory is associated with a detection, the TDAM observes a new appearance and its parameters are incrementally updated.

%Suppose that the TDAM $\bm{\theta}_t$ observes a new appearance $\mathbf{o}_{t+1}$ at time $(t+1)$, the task is to adapt $\bm{\theta}_t$ to the new appearance $\mathbf{o}_{t+1}$, resulting in a new TDAM $\bm{\theta}_{t+1}$. Rather than considering $\mathbf{o}_{t+1}$ independently, we use the subsequence $\mathbf{W}_{t+1}$ as the newly obtained training data. Note that the subsequence $\mathbf{W}_{t+1}$ can be acquired by incorporating $\mathbf{o}_{t+1}$ into $\mathbf{W}_{t}$, where the earliest appearance is removed to ensure the fixed-length. Without introducing ambiguity, we use the consistent form to denote the subsequence $\mathbf{W}_{t+1} = \{ \mathbf{o}_{t_1},\ldots,\mathbf{o}_{t_L} \}$.

%Given the current model $\bm{\theta}_t = \left(\bm{\pi}^{(t)}, \mathbf{A}^{(t)}, \mathbf{F}^{(t)}\right)$ and a newly obtained training sequence $\mathbf{W}_{t+1}$, we utilize an online Expectation-Maximization (EM) algorithm \cite{digalakis1999online} for TDAM parameters updating.

Suppose that the TDAM $\bm{\theta}_t$ observes a new appearance at time $(t+1)$, the task is to update the parameters $\bm{\theta}_t$ to $\bm{\theta}_{t+1}$ by considering the newly obtained subsequence $\mathbf{W}_{t+1} = \{ \mathbf{o}^{(t+1)}_{t_1},\ldots,\mathbf{o}^{(t+1)}_{t_L} \}$ as a training sequence, as shown in Fig.~\ref{Matching-Example}(d). Note that the subsequence $\mathbf{W}_{t+1}$ can be acquired by incorporating the new appearance into $\mathbf{W}_{t}$, where the earliest appearance is removed to ensure the fixed-length. With the current model and a training sequence, we utilize an efficient online Expectation-Maximization (EM) algorithm \cite{digalakis1999online} for TDAM parameters updating.

At the \emph{E-step}, we apply a forward-backward procedure \cite{rabiner1989tutorial} on the training sequence $\mathbf{W}_{t+1}$ using the current model $\bm{\theta}_t$ to compute the following probabilities,
\begin{align}
    \xi^{(t)}_l(i,j) & = P\left( s_{t_l} = i, s_{t_{l+1}} = j | \mathbf{o}_{t_1:t_L},\bm{\theta}_t  \right), \label{eq:4} \\
    \gamma^{(t)}_l(i,k) & = P\left( s_{t_l} = i, \delta_l(s_{t_l}) = k | \mathbf{o}_{t_1:t_L},\bm{\theta}_t  \right), \label{eq:5}
\end{align}
where $i = 1,\ldots,N$, $j = 1,\ldots,N$, $k = 1,\ldots,M$, and $\delta_l(s_{t_l})$ indicates the mixture component for state $s_{t_l}$ at time $t_l$. Note that we drop the time index of the appearances within $\mathbf{W}_{t+1}$ for simplicity, and denote the sequence as $\mathbf{o}_{t_1:t_L}$. To be specific, $\xi^{(t)}_l(i,j)$ represents the probability of being in state $i$ at time $t_l$, and state $j$ at time $t_{l+1}$, given the model $\bm{\theta}_t$ and the sequence $\mathbf{o}_{t_1:t_L}$. And, $\gamma^{(t)}_l(i,k)$ is the probability of being in state $i$ at time $t_l$ with the $k$-th mixture component accounting for the appearance $\mathbf{o}_{t_l}$.

Then we have the following expected sufficient statistics:
\begin{align}
    \xi^{(t)}_{ij} &= \sum\nolimits_{l=1}^{L-1} \xi^{(t)}_l(i,j), \label{eq:6}\\
    \gamma^{(t)}_{ik} &= \sum\nolimits_{l=1}^{L} \gamma^{(t)}_l(i,k), \label{eq:7}\\
    \mathbf{m}^{(t)}_{ik} &= \sum\nolimits_{l=1}^{L} \gamma^{(t)}_l(i,k) \cdot \mathbf{o}_{t_l}, \label{eq:8}\\
    C^{(t)}_{ik} &= \sum\nolimits_{l=1}^{L} \gamma^{(t)}_l(i,k) \cdot \mathbf{o}_{t_l}\mathbf{o}_{t_l}^{\top}. \label{eq:9}
\end{align}
To account for the previously-acquired knowledge, we accumulate the expected sufficient statistics (stating from zero) over time. Denote the accumulated expected sufficient statistics at time $t$ as $\hat{\xi}^{(t)}_{ij}$, $\hat{\gamma}^{(t)}_{ik}$, $\hat{\mathbf{m}}^{(t)}_{ik}$, and $\hat{C}^{(t)}_{ik}$, respectively. The accumulated expected sufficient statistics at time $(t+1)$ are calculated as
\begin{align}
    \hat{\xi}^{(t+1)}_{ij} &= (1-\eta) \cdot \hat{\xi}^{(t)}_{ij} + \eta \cdot \xi^{(t)}_{ij}, \label{eq:10}\\
    \hat{\gamma}^{(t+1)}_{ik} &= (1-\eta) \cdot \hat{\gamma}^{(t)}_{ik} + \eta \cdot \gamma^{(t)}_{ik}, \label{eq:11}\\
    \hat{\mathbf{m}}^{(t+1)}_{ik} &= (1-\eta) \cdot \hat{\mathbf{m}}^{(t)}_{ik} + \eta \cdot \mathbf{m}^{(t)}_{ik}, \label{eq:12}\\
    \hat{C}^{(t+1)}_{ik} &= (1-\eta) \cdot \hat{C}^{(t)}_{ik} + \eta \cdot C^{(t)}_{ik}, \label{eq:13}
\end{align}
where $\eta$ is the learning rate. We set $\eta = 0.8$ in our implementation.

At the \emph{M-step}, we use the accumulated expected sufficient statistics at time $(t+1)$ to estimate the new parameters $\bm{\theta}_{t+1} = \left(\bm{\pi}^{(t+1)}, \mathbf{A}^{(t+1)}, \mathbf{F}^{(t+1)}\right)$, expressed as
\begin{align}
    a^{(t+1)}_{ij} &= \frac{\hat{\xi}^{(t+1)}_{ij}}{\sum_{j=1}^{N}\hat{\xi}^{(t+1)}_{ij}}, \label{eq:14}\\
    \omega^{(t+1)}_{ik} &= \frac{\hat{\gamma}^{(t+1)}_{ik}}{\sum_{k=1}^{M}\hat{\gamma}^{(t+1)}_{ik}}, \label{eq:15}\\
    \bm{\mu}^{(t+1)}_{ik} &= \frac{1}{\hat{\gamma}^{(t+1)}_{ik}} \cdot \hat{\mathbf{m}}^{(t+1)}_{ik}, \label{eq:16}\\
    \Sigma^{(t+1)}_{ik} &= \frac{1}{\hat{\gamma}^{(t+1)}_{ik}} \cdot \hat{C}^{(t+1)}_{ik} - \left[\bm{\mu}^{(t+1)}_{ik}\right]\left[\bm{\mu}^{(t+1)}_{ik}\right]^{\top}. \label{eq:17}
\end{align}
Note that the initial state distribution is not updated online, \ie, $\bm{\pi}^{(t)}$ is fixed to be uniform distribution. The reason is that the subsequences used for online learning might start at any time, thus we do not apply any prior information to the initial states.

\textbf{Initialization:} The online EM algorithm can be sensitive to the initialization of the model parameters. In this work, we learn a general TDAM that roughly characters the spatial and temporal properties of human appearances beforehand. Specifically, we first perform K-means clustering on the static human appearances from the INRIA dataset in the learned mid-level feature space to form $N$ clusters. The resulting $N$ clusters are used to estimate the observation densities of $N$ hidden states for the general TDAM. $M$ weighted Gaussian distributions are estimated to optimally represent each cluster. In terms of initial state distribution and state transition matrix, we simply apply uniform distributions. For clarity, we denote the general TDAM as $\bm{\theta}_0 = \left(\bm{\pi}^{(0)}, \mathbf{A}^{(0)}, \mathbf{F}^{(0)}\right)$. The person-specific TDAMs for all tracked persons are learned incrementally from a same initialization $\bm{\theta}_0$ by incorporating the person-specific properties of appearance sequences.

\section{Online Multi-Person Tracking}
\label{MOT}

In this section, we present our online multiple people tracking method that uses the TDAM to improve tracking performance. The method operates entirely in the image coordinate without camera or ground plane calibration. We adopt a popular data association framework \cite{wu2007detection,xing2009multi,shu2012part,BaeY2014robust} to formulate the online multi-person tracking problem. At each frame, the detections are linked with existing trajectories by solving a data association problem, and then the trajectories are sequentially grown with the associated detections. Trajectory initialization and termination strategies are also involved to make the tracking method complete. Note that more sophisticated formulation can be involved to further improve the tracker. In this work, however, we embed the TDAM into a simple yet efficient formulation to highlight its importance and practical usefulness in multi-person tracking.

\subsection{Data Association}
\label{DA}

Suppose that we have $n$ trajectories $\mathbb{X}_{t} = \{X^p_{t}\}_{p=1}^{n}$ at time $t$ and $m$ detections $\mathbb{Z}_{t+1} = \{\mathbf{z}^q_{t+1}\}_{q=1}^{m}$ at time $t+1$, where $X^p_t$ is the trajectory of the $p$-th person and $\mathbf{z}^q_{t+1}$ is the $q$-th detection provided by the detector. To perform association between $\mathbb{X}_{t}$ and $\mathbb{Z}_{t+1}$, an $n \times m$ association cost matrix $\Psi$ is computed by
\begin{equation}\label{eq:18}
    \Psi_{pq}=-\log\left(\rho(X^p_{t},\mathbf{z}^q_{t+1})\right),
\end{equation}
where $\rho(X^p_{t},\mathbf{z}^q_{t+1})$ is the affinity between the trajectory $X^p_{t}$ and the detection $\mathbf{z}^q_{t+1}$. Then the data association problem is solved by using the Hungarian algorithm \cite{kuhn1955hungarian} such that the total cost in $\Psi$ is minimized. When the association cost of a trajectory-detection pair is less than a pre-defined threshold $-\log(\epsilon)$, the detection $\mathbf{z}^q_{t+1}$ is associated with $X^p_t$. A Kalman filter is used to refine the object states of a trajectory, with the associated detections as the measurement data.

We then present the details of computing the affinities $\rho(X^p_{t},\mathbf{z}^q_{t+1})$. A trajectory at time $t$ is represented as $X^p_{t} = \{\bm{\theta}^p_t, M^p_t, S^p_t \}$, where $\bm{\theta}^p_t$ is the online-learned TDAM, and $M^p_t$ and $S^P_t$ indicate the motion and shape information, respectively. Similarly, we describe a detection as $\mathbf{z}^q_{t+1} = \{ \mathbf{o}^q_{t+1},\mathbf{c}^q_{t+1},\mathbf{s}^q_{t+1} \}$, where $\mathbf{o}^q_{t+1}$, $\mathbf{c}^q_{t+1}$, and $\mathbf{s}^q_{t+1}$ are, respectively, the appearance, position, and size of the detection response. The affinity between $X^p_{t}$ and $\mathbf{z}^q_{t+1}$ can be calculated based on the appearance, motion, and shape affinities, expressed as
\begin{eqnarray}\label{eq:19}
\begin{split}
    &\rho(X^p_{t},\mathbf{z}^q_{t+1}) = \rho_A(X^p_{t},\mathbf{z}^q_{t+1}) \cdot \rho_M(X^p_{t},\mathbf{z}^q_{t+1}) \cdot \rho_S(X^p_{t},\mathbf{z}^q_{t+1}).
\end{split}
\end{eqnarray}

\textbf{Appearance Model:} Given an image patch of one target person, we compute the HOG feature and project it into a mid-level feature space with dimensionality $d$ (as described in Sec.~\ref{pretrain}) to achieve a semantic interpretation. For each traced person, we build a specific TDAM to model both the spatial structure and the temporal dynamic of the varying appearances. With the TDAM, the appearance affinity between $X^p_{t}$ and $\mathbf{z}^q_{t+1}$ is given by
\begin{eqnarray}\label{eq:20}
\begin{split}
    \rho_A(X^p_{t},\mathbf{z}^q_{t+1}) = P\left( \mathbf{o}^q_{t+1} | \bm{\theta}^p_t \right),
\end{split}
\end{eqnarray}
which is defined by Eq.~(\ref{eq:1}) and Eq.~(\ref{eq:2}).

%We integrate several cues including color, shape and texture to represent the appearances of persons. Each image patch of person is normalized to $64\times32$ pixels. For color information, we use RGB color histograms with $64$-bin for each channel, resulting a $192$-dimensional vector. To describe shape information, we use the HOG feature \cite{dalal2005histograms} with $8\times8$ cell and $36$-dimensional gradient histogram for each cell. The final shape feature has $1152$ dimensions. To capture texture information, we use the LBP histogram \cite{ojala1996comparative} with.

\textbf{Motion Model:} We use a constant velocity motion model to predict the person position for a trajectory. The difference between the predicted position of a trajectory and the position of a detection is assumed to follow a Gaussian distribution with covariance $\Lambda$. Thus we evaluate the motion affinity between $X^p_{t}$ and $\mathbf{z}^q_{t+1}$ as
\begin{eqnarray} \label{eq:21}
\begin{split}
    \rho_M(X^p_{t},\mathbf{z}^q_{t+1}) = \mathcal{N}\left( \mathbf{c}_{tail}^p + \tau\mathbf{v}^p;\mathbf{c}^q_{t+1}, \Lambda \right),
\end{split}
\end{eqnarray}
where $\mathbf{c}_{tail}^p$ is the last refined position of the trajectory $X^p_t$, $\mathbf{v}^p$ is the velocity of the person estimated by Kalman filter, and $\tau$ is the frame gap between $\mathbf{c}_{tail}^p$ and $\mathbf{c}^q_{t+1}$.

\textbf{Shape Model:} The shape affinity between $X^p_{t}$ and $\mathbf{z}^q_{t+1}$ is evaluated with their heights and widths. Denote the size of the detection $\mathbf{z}^q_{t+1}$ as $\mathbf{s}^q_{t+1} = \{h^q,w^q\}$, where $h^q$ and $w^q$ are the height and width of the detection, respectively. The shape information of the trajectory $X^p_{t}$ can be represented as $S^p_t = \{h_X^p,w_X^p\}$, where $h_X^p$ and $w_X^p$ are the average height and width of the person, respectively. Then the shape affinity is evaluated as
\begin{eqnarray} \label{eq:22}
\begin{split}
    \rho_S(X^p_{t},\mathbf{z}^q_{t+1}) = \exp\left( -\frac{1}{2}\left(\frac{|h_X^p-h^q|}{h_X^p+h^q} + \frac{|w_X^p-w^q|}{w_X^p+w^q} \right) \right).
\end{split}
\end{eqnarray}

\subsection{Trajectory Initialization and Termination}

In order to find new trajectories, we link the detections that are not already associated with any existing trajectories to form short tracklets. When the length of a tracklet grows over the threshold $T_{init}$, we generate a new trajectory with a TDAM with initial parameters $\bm{\theta}_{0}$. Then the appearance sequence from the short tracklet is used to adapt the TDAM to the newly tracked person, using the incremental learning algorithm presented in Sec.~\ref{incremental}. For trajectory termination, we simply cancel the trajectories that have no associated detections for $T_{term}$ subsequent frames. In addition, trajectories are terminated if the corresponding persons exit the field-of-view. In our implementation, we set $T_{init} = 5$, and $T_{term} = 5$.

\section{Experiments}

In this section, we perform an extensive experimental evaluation on various datasets to validate the effectiveness and the superiority of our method. After the detailed description of datasets, evaluation metrics, and parameter settings, we analyze our method in two regards. First, we examine the influence of the carefully designed components involved in our method on the tracking performance (diagnosis analysis). Next, we compare our method with a number of state-of-the-art algorithms in a way that is strictly fair (quantitative comparison). The runtime performance of our method is also discussed in the following.

\subsection{Datasets}

We use the public available MOTChallenge 2015 Benchmark \cite{leal2015motchallenge} for experimental evaluation, which gathers various existing and new challenging video sequences. Since our method performs tracking on the image coordinate, we use the 2D MOT 2015 sequences in the MOTChallenge 2015. The sequences are composed of $11$ training and $11$ testing video sequences in which the challenges include camera motion, low viewpoint, varying frame rates, and server weather condition. The training sequences contain over $5500$ frames ($\sim 7$ minutes) and $500$ annotated trajectories ($39905$ bounding boxes). The benchmark releases the ground truth of the training sequences publicly, and thus we use the training sequences to determine the set of system parameters and for diagnosis analysis. The testing sequences contains over $5700$ frames ($\sim 10$ minutes) and $721$ annotated trajectories ($61440$ bounding boxes), while the annotations are not available to avoid (over)fitting of the competing methods to the specific sequences. Since it is hard for methods to overtune on such a large amount of data, we use the testing sequences for quantitative comparison against various state-of-the-art trackers. Moreover, the tracking results of all competing methods are automatically evaluated by the benchmark, and the performance scores are public online for strictly fair comparison.

\subsection{Evaluation Metrics}

We use the widely accepted CLEAR performance metrics \cite{keni2008evaluating} for quantitative evaluation: the multiple object tracking precision (MOTP$\uparrow$) that evaluates average overlap rate between true positive tracking results and the ground truth, and the multiple object tracking accuracy (MOTA$\uparrow$) which indicates the accuracy composed of false positives (FP$\downarrow$), false negatives (FN$\downarrow$) and identity switches (IDS$\downarrow$). Additionally, we report measures defined by Li \etal \cite{li2009learning}, including the percentage of mostly tracked (MT$\uparrow$) and mostly lost (ML$\downarrow$) ground truth trajectories, as well as the number of times that a ground truth trajectory is interrupted (Frag$\downarrow$). The false positive ratio is also measured by the number of false alarms per frame (FAF$\downarrow$). Here, $\uparrow$ means that higher scores indicate better results, and $\downarrow$ represents that lower is better.

For fair comparison, we use the public detections for all sequences as well as the performance evaluation script provided by the MOTChallenge 2015. Note that the detections are generated by using the recent object detector \cite{dollar2014fast} trained on the INRIA dataset \cite{dalal2005histograms} with default parameters.

\begin{figure*}[t]
    \centering
    \subfigure[]{
    \includegraphics[width=0.31\textwidth]{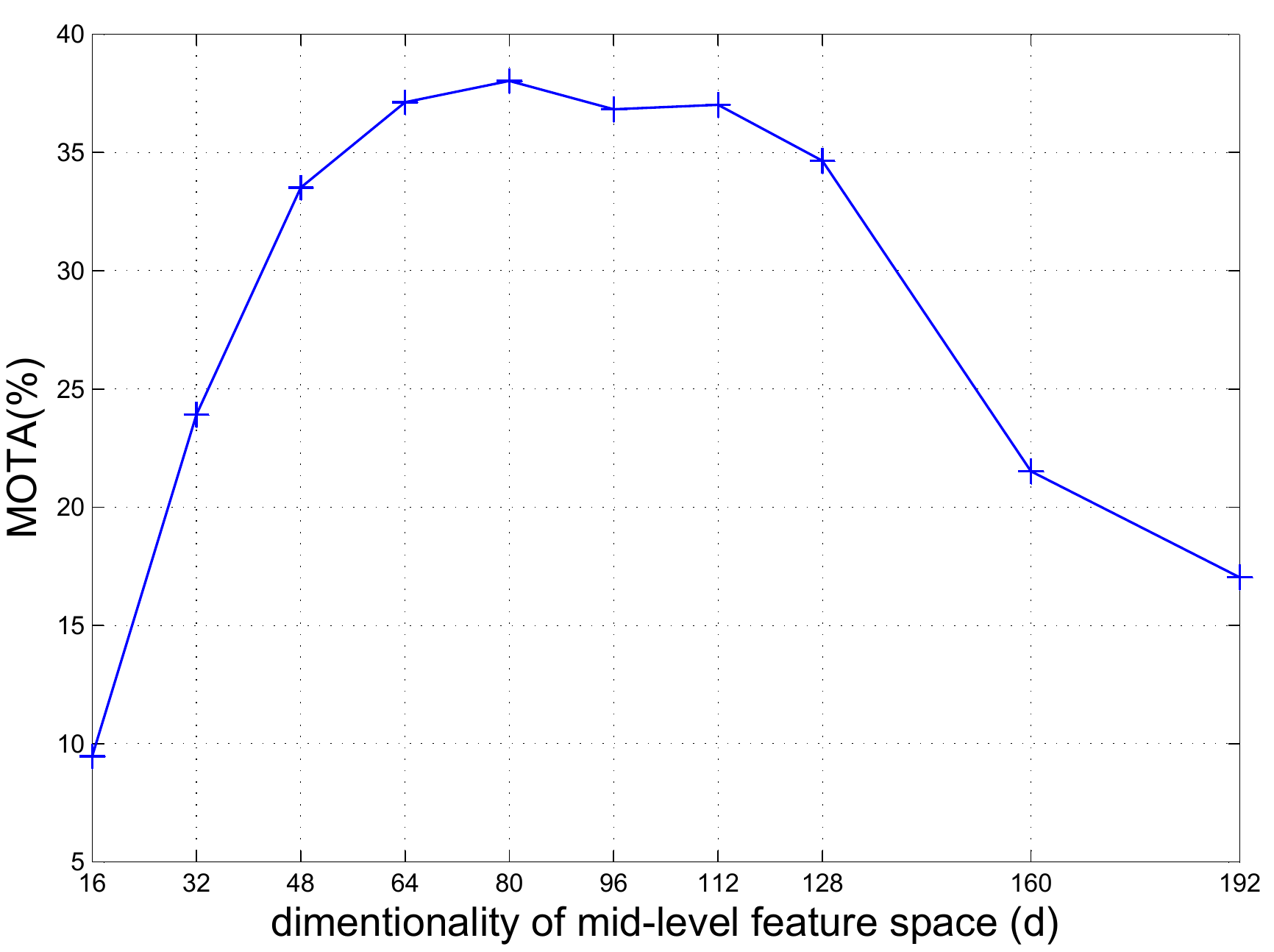}}
    \subfigure[]{
    \includegraphics[width=0.31\textwidth]{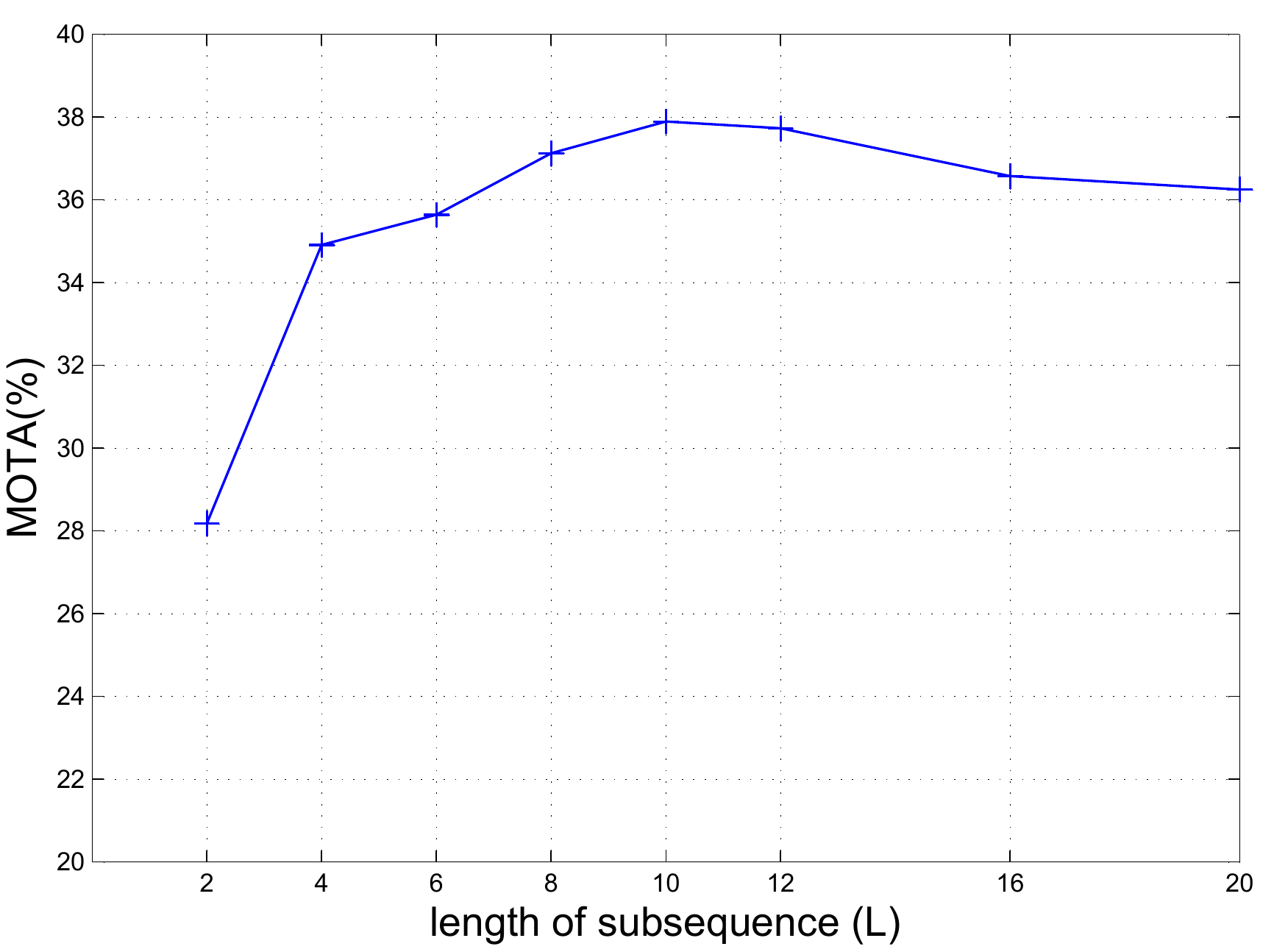}}
    \subfigure[]{
    \includegraphics[width=0.31\textwidth]{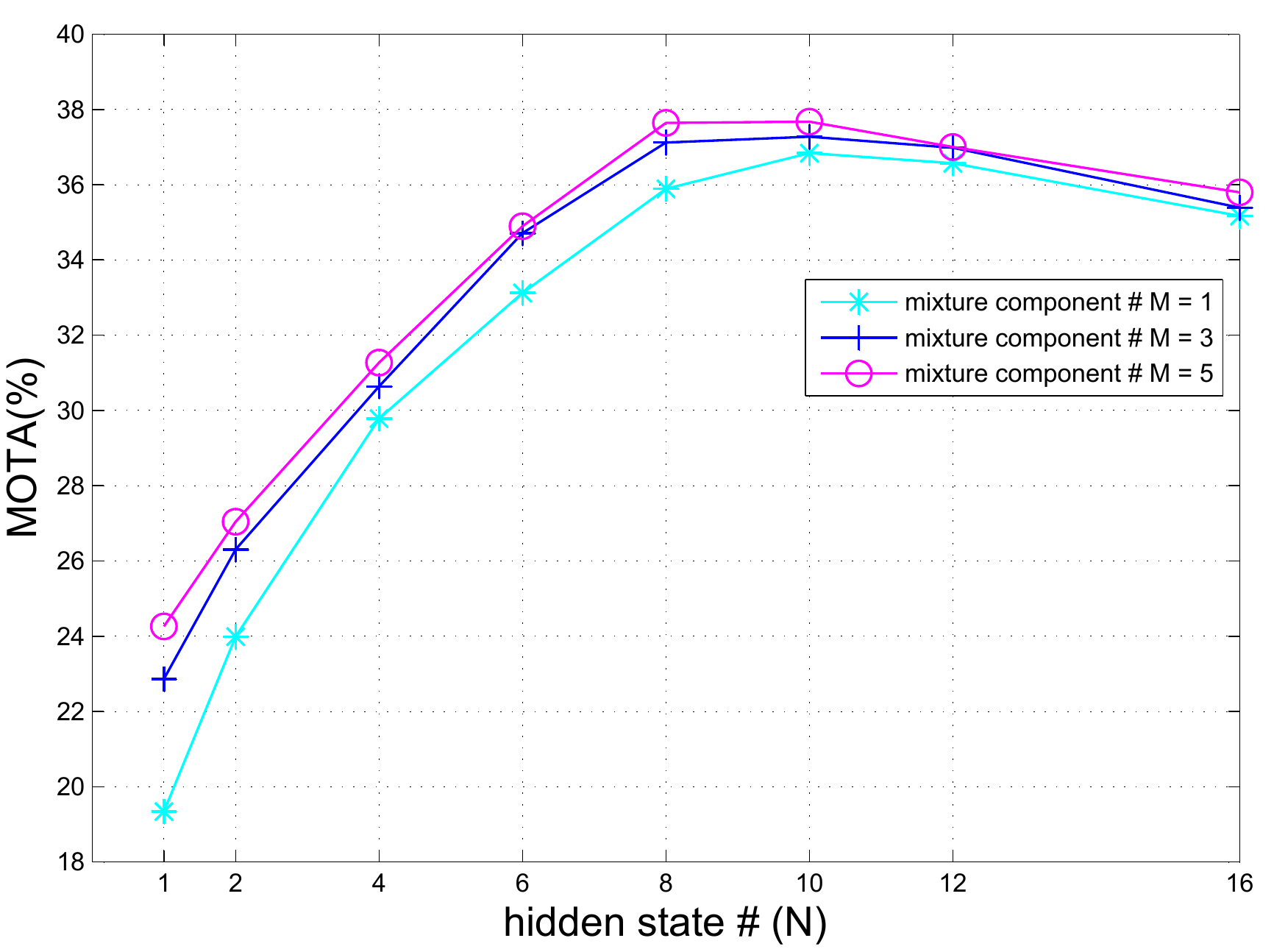}}
    \caption{Influence of individual parameters on tracking performance. Four critical parameters are experimentally studied: (a) the dimensionality $d$ of mid-level feature space; (b) the length $L$ of subsequence $\mathbf{W}_t$; (c) the number of hidden states $N$ and the number of mixture components $M$. Each plot shows the relative change in performance (measured by the average MOTA on all training sequences) by changing the value of a single parameter while keeping the other ones fixed. As can be seen, the results are stable over a range of settings.}
    \vspace{-0pt}
    \label{ParamSetting}
\end{figure*}

\subsection{Parameter Settings}

All parameters involved in our method are found experimentally using the training sequences, and remain fixed throughout the experiments. The dimensionality of the mid-level feature space (\ie, the number of detectors selected by the feature selection approach) is set to $d = 64$. The number of hidden states $N$ and the number of mixture components $M$ defined in the TDAM are set to $N = 8$ and $M = 3$, respectively. To perform incremental learning of the TDAM, we set the length of subsequence $\mathbf{W}_t$ as $L = 8$, which has been verified to be sufficient for the TDAM to capture the apparent appearance change between two consecutive frames. To compute the affinity using Eq.~(\ref{eq:19}), the only one parameter need to be set beforehand is the covariance $\Lambda$ in Eq.~(\ref{eq:21}), which is set to $\Lambda = diag(20^2,50^2)$ in our experiment.

To further study the influence of the critical parameters, we run our tracking method repetitively on the $11$ training sequences and modify the corresponding parameter while keeping all the other ones fixed. The relative change in performance, measured by the average MOTA on all training sequences, is plotted against the parameter value in Fig.~\ref{ParamSetting}. As can be seen, the results are stable over a range of settings. The strong decline only can be observed when the parameters are set to be unreasonable. For example, as shown in Fig.~\ref{ParamSetting}(a), the MOTA score drastically decrease when we choose a extremely small/large value for the dimensionality $d$ of the mid-level feature space. The reason is that the dynamic changes of human appearance are hardly captured by a small set of mid-level features. On the other hand, a large amount of features increase the redundancy and the model complexity, and thus degrade the TDAM. Similar results are observed for other parameters in Fig.~\ref{ParamSetting}(b) and  Fig.~\ref{ParamSetting}(c). We can conclude that the tracking performance of our method is only slightly affected by parameter changes within a reasonable range. Moreover, our choice of parameters is rather conservative and does not correspond to the best setting. This indicates that the model has not been overtuned on the training data.

\begin{table*}[t]
\caption{Average tracking performance of our method (p1) and four baseline algorithms (a1)-(a4) on the training sequences of the MOTChallenge 2015. The results are listed by two groups for better readability: quantitative comparisons between (a1), (a2), and (p1) highlight the superiority of the TDAM, and quantitative comparisons between (a3), (a4), and (p1) demonstrate the effectiveness of the mid-level semantic features.}
\vspace{-0pt}
\label{diagnosis}
\begin{center}
\newcolumntype{X}{p{15mm}<{\centering}}
\newcolumntype{Y}{p{11mm}<{\centering}}
\newcolumntype{Z}{p{8mm}<{\centering}}
\begin{tabular}{|p{34mm}<{\raggedright}|XXZYYYYZZ|}
\hline
\textbf{Method} & \textbf{MOTA}[\%]$\uparrow$ & \textbf{MOTP}[\%]$\uparrow$ & \textbf{FAF}$\downarrow$ & \textbf{MT}[\%]$\uparrow$ & \textbf{ML}[\%]$\downarrow$ & \textbf{FP}$\downarrow$ & \textbf{FN}$\downarrow$ & \textbf{IDS}$\downarrow$ & \textbf{FG}$\downarrow$ \\
\hline\hline
(a1): baseline algorithm
&       {24.2$\pm$27.3}&       {72.8}&       { 1.7}&       {17.0}&       {47.8}&       { 9,325}&       {20,393}&       {  526}&     {1,327}  \\
(a2): TDAM w/o temp.
&       {29.0$\pm$25.1}&       {73.0}&       { 1.6}&       {19.6}&       {44.4}&       { 8,789}&       {19,252}&       {  294}&     {  910}  \\
(p1): TDAM
&       {37.1$\pm$23.8}&       {73.4}&       { 1.2}&       {20.4}&       {42.4}&       { 6,694}&       {18,213}&       {  184}&     {  675}  \\
\hline\hline
(a3): TDAM with PCA
&       {30.0$\pm$21.7}&       {73.2}&       { 1.4}&       {20.6}&       {46.8}&       { 7,882}&       {19,911}&       {  158}&     {  738}  \\
(a4): TDAM with LLE
&       {29.0$\pm$22.9}&       {73.4}&       { 1.5}&       {21.0}&       {44.8}&       { 8,579}&       {19,575}&       {  185}&     {  772}  \\
(p1): TDAM
&       {37.1$\pm$23.8}&       {73.4}&       { 1.2}&       {20.4}&       {42.4}&       { 6,694}&       {18,213}&       {  184}&     {  675}  \\
\hline
\end{tabular}
\end{center}
\end{table*}

\begin{table*}[t]
\caption{Quantitative evaluation accuracy of our method (TDAM) and other state-of-the-art approaches on the MOTChallenge test video sequences. Result listings are grouped into online and offline methods. \txtred{Bold} scores highlight the best results of online methods, and \txtblu{blue} scores indicates the best results of offline methods.}
\vspace{-0pt}
\label{quatitative}
\begin{center}
\newcolumntype{X}{p{15mm}<{\centering}}
\newcolumntype{Y}{p{11mm}<{\centering}}
\newcolumntype{Z}{p{8mm}<{\centering}}
\begin{tabular}{p{3mm}<{\centering}|p{26.5mm}<{\raggedright}|XXZYYYYZZ|}
\hline
&\textbf{Method} & \textbf{MOTA}[\%]$\uparrow$ & \textbf{MOTP}[\%]$\uparrow$ & \textbf{FAF}$\downarrow$ & \textbf{MT}[\%]$\uparrow$ & \textbf{ML}[\%]$\downarrow$ & \textbf{FP}$\downarrow$ & \textbf{FN}$\downarrow$ & \textbf{IDS}$\downarrow$ & \textbf{FG}$\downarrow$ \\
\hline\hline
&TC\_ODAL~\cite{BaeY2014robust}
&       {15.1$\pm$15.0}&       {70.5}&       { 2.2}&       { 3.2}&       {55.8}&       {12,970}&       {38,538}&       {  637}&       {1,716}  \\
&RMOT~\cite{yoon2015bayesian}
&       {18.6$\pm$17.5}&       {69.6}&       { 2.2}&       { 5.3}&       {53.3}&       {12,473}&       {36,835}&       {  684}&\txtred{1,282}  \\
\rowcolor{mygray}
\multirow{-3}{*}{\begin{sideways}{online}\end{sideways}}
&TDAM
&\txtred{33.0$\pm$9.8}&\txtred{72.8}&\txtred{ 1.7}&\txtred{13.3}&\txtred{39.1}&\txtred{10,065}&\txtred{30,617}&\txtred{  464}&       {1,506}  \\
\hline
\hline
\multirow{6}{*}{\begin{sideways}{offline}\end{sideways}}
&SegTrack~\cite{milan2015joint}
&       {22.5$\pm$15.2}&\txtblu{71.7}&\txtblu{ 1.4}&       { 5.8}&       {63.9}&\txtblu{ 7,890}&       {39,020}&\txtblu{  697}&\txtblu{  737}  \\
&MotiCon~\cite{leal2014learning}
&\txtblu{23.1$\pm$16.4}&       {70.9}&       { 1.8}&       { 4.7}&       {52.0}&       {10,404}&       {35,844}&       {1,018}&       {1,061}  \\
&CEM~\cite{Milan:2014:CEM}
&       {19.3$\pm$17.5}&       {70.7}&       { 2.5}&\txtblu{ 8.5}&       {46.5}&       {14,180}&\txtblu{34,591}&       {  813}&       {1,023}  \\
&TBD~\cite{geiger20143d}
&       {15.9$\pm$17.6}&       {70.9}&       { 2.6}&       { 6.4}&       {47.9}&       {14,943}&       {34,777}&       {1,939}&       {1,963}  \\
&SMOT~\cite{dicle2013way}
&       {18.2$\pm$10.3}&       {71.2}&       { 1.5}&       { 2.8}&       {54.8}&       { 8,780}&       {40,310}&       {1,148}&       {2,132}  \\
&DP\_NMS~\cite{pirsiavash2011globally}
&       {14.5$\pm$13.9}&       {70.8}&       { 2.3}&       { 6.0}&\txtblu{40.8}&       {13,171}&       {34,814}&       {4,537}&       {3,090}  \\
\hline
\end{tabular}
\end{center}
\end{table*}

\subsection{Diagnosis Analysis}

As previously mentioned, we address the problem of online multi-person tracking by using a temporal dynamic appearance modeling algorithm. The temporal dynamic characteristics, as well as the spatial structure, of varying appearances are explicitly modeled by the TDAM to provide more accurate appearance affinities to guide data association. Moreover, we also propose a feature selection approach to find an appropriate mid-level semantic feature space for temporal dynamic appearance modeling. The resulting semantic interpretation of appearance changes benefits the capture of temporal dynamic. To demonstrate the effectiveness of our design, we build several baseline algorithms to do validation and analyze various aspects of our approach.

We first build two baseline algorithms (a1) and (a2) to investigate the effect of the TDAM in terms of multi-person tracking. The algorithm (a1) is the baseline algorithm simply implemented by using the distances between the mid-level features to compute appearance affinities. The algorithm (a2) is a simplified version of the TDAM which is obtained by removing the temporal dependencies, resulting in a mixture densities model that only describes the spatial structure of dynamic appearances. For fair comparison, we set the number of mixture components in the algorithm (a2) as $N \times M$ to be consistent with the full implementation of the TDAM. The comparison results between our method, denoted as (p1), and these two baseline algorithms on the training sequences of the MOTChallenge 2015 are listed in the first group in Table \ref{diagnosis}. The algorithm (a1) discards both the spatial structure and the temporal dynamic properties of human appearances, and thus shows severe performance degradation in terms of both tracking accuracy and consistency. While the algorithm (a2) results in apparent performance improvement compared to the algorithm (a1). It validates that using mixture densities to model the spatial structure of appearance variations is effective. Our method incorporates temporal dynamic to increase the discriminativeness of affinity measurement by using the TDAM, and thus show the best performance. As can be observed from the comparison results, our method produces significantly better results in terms of MOTA, MT and ML, and apparently reduces the number of FP, FN and IDS.

Although the TDAM has been proved to be useful in terms of multi-person tracking, the feature space in which the TDAM features its discriminative ability is critical. We build another two baseline algorithms (a3) and (a4) to validate that the feature selection approach proposed in this paper facilitates the modeling of temporal dynamic. The algorithm (a3) employs Principal Component Analysis (PCA) to find a low-dimensional linear subspace for the TDAM, and the algorithm (a4) employs the Locally Linear Embedding (LLE) \cite{roweis2000nonlinear} to seek a nonlinear appearance manifold. Both (a3) and (a4) use the same training data as our feature selection approach as described in Sec.~\ref{pretrain}, and map the low-level HOG feature into a $d$-dimensional feature space. The comparison results between our method and these two baseline algorithms on the training sequences of the MOTChallenge 2015 are listed in the second group in Table \ref{diagnosis}. As can be seen, our method (p1) achieves apparently better overall performance than (a3) and (a4), especially reduces the number of FP and FN. This is an indication that the mid-level semantic feature space used in our method performs well in terms of characterizing the temporal dynamic properties of appearance changes. Compared with the common used manifold learning approaches, the proposed feature selection approach provides richer semantic interpretation of human appearances and thus demonstrates its usefulness in the TDAM.

\subsection{Quantitative Comparison}

Quantitative evaluation results
\footnote{The comparison is also available at the website of the MOTChallenge \url{http://motchallenge.net/results/2D_MOT_2015/}.}
of our algorithm compared with the state-of-the-art tracking methods on the testing sequences of the MOTChallenge 2015 are listed in Table~\ref{quatitative}. The state-of-the-art trackers include TC\_ODAL~\cite{BaeY2014robust}, RMOT~\cite{yoon2015bayesian}, SegTrack~\cite{milan2015joint}, MotiCon~\cite{leal2014learning}, CEM~\cite{Milan:2014:CEM}, TBD~\cite{geiger20143d}, SMOT~\cite{dicle2013way}, and DP\_NMS~\cite{pirsiavash2011globally}, in which the TC\_ODAL and RMOT trackers are online algorithms while the other trackers perform multi-person tracking in a batch mode (offline).

Overall, our method outperforms other state-of-the-art trackers with a significant margin ($\sim 10\%$), even compared to offline methods. Our methods achieves the lowest identity switch (IDS) while achieving the highest detection accuracy (lowest FP and FN) and tracking consistency (highest MT and lowest ML). In turn, our method records the highest MOTA, which is a good approximation of the overall performance. It owes to the TDAM that produces accurate and reliable appearance affinities between trajectories and detections. By exploiting the temporal dynamic characteristics contained in appearance sequences, the TDAM is able to provide helpful information to achieve robust association, even when the appearances originated from different trajectories turn to be similar at some static times. The quantitative evaluation results demonstrate that the trajectories produced by our method are more consistent under various challenging conditions.

\begin{figure*}
    \centering
    \includegraphics[width=0.98\textwidth]{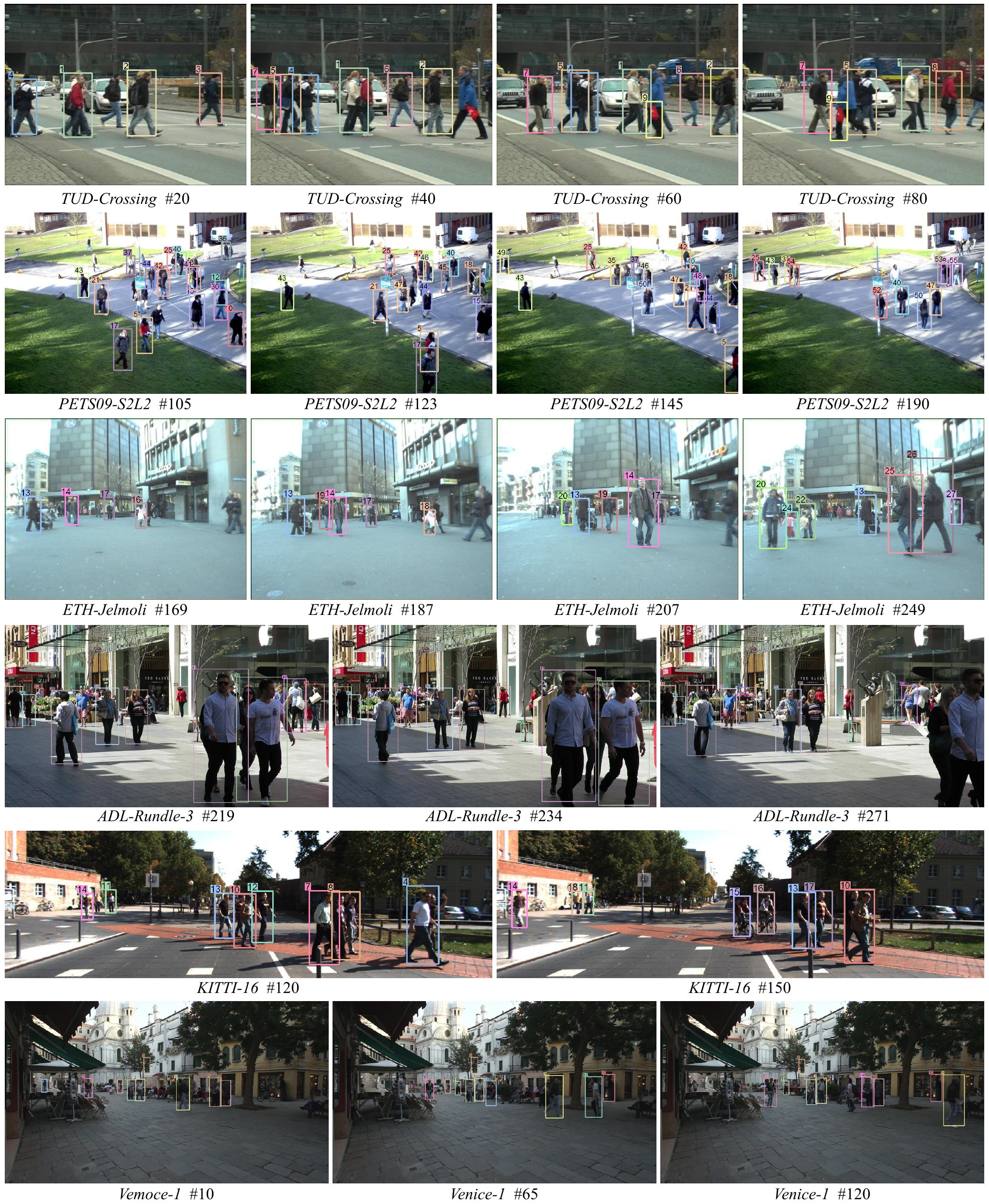}
    \vspace{-0pt}
    \caption{Sample tracking results of our method on six representative testing video sequences of the MOTChallenge 2015 (\ie, \emph{TUD-Crossing} \emph{PETS09-S2L2}, \emph{ETH-Jelmoli}, \emph{ADL-Rundle-3}, \emph{KITTI-16}, and \emph{Venice-1}). At each frame, persons with different IDs are indicated by bounding boxes with different colors. Best viewed in color. (Refer to the tracking videos for more detailed results.)}
    \label{sampleresults}
\end{figure*}

Several qualitative examples of tracking results produced by our method on the testing sequences are shown in Fig.~\ref{sampleresults}. Consistency of the estimated trajectories is indicated by bounding boxes of the same color on the same object over time. Our method is able to accurately track the target persons against the inference of similar-looking objects, short-term occlusions, abundant false positive detections etc. (Videos suitable for qualitative evaluation of the results across all frames are available at \url{http://iitlab.bit.edu.cn/mcislab/~yangmin}. The detailed tracking results provided by our method and the state-of-the-art algorithms on the testing sequences could be found in the website of the MOTChallenge \url{http://motchallenge.net/results/2D_MOT_2015/}.)

On the other hand, our method produces slightly more fragmented trajectories in return. The reason is that our method only exploits the observations up to the current frame to sequentially build the trajectories, and might be affected by long term occlusions and abrupt motions. Since the temporal consistency is interrupted in these situations, the TDAM tends to terminate the trajectory and causes the increasing of the FG score. While the IDS score are significantly reduced due to the strong discriminative ability of the TDAM. In addition, our method might be confused by consistent false alarms in background clutters (see results on the sequences \emph{ADL-Rundle-3} and \emph{Venice-1} in Fig.~\ref{sampleresults} for examples). The offline approaches consider both the previous and future frames to associate detections, and produce longer trajectories and much fewer false positive in contrast. These cases can be handled by extended the TDAM to consider a part of future frames with a short-term latency in return.

\subsection{Speed}

Our method runs at about 5 fps on an Intel Core i7 3.5 GHz PC with 16 GB memory, and the MATLAB code does not involve optimization. The runtime relies on the number of trajectories. The most time-consuming part is the parameters update for the TDAM (accounting for $50\%$ of the total computation for each frame). We can significantly reduce the runtime by simultaneously updating the parameters for all TDAMs using parallel programming, since the incremental learning for each person-specific TDAM is independent after data association.

\section{Conclusions}

In this paper, we have proposed a novel appearance modeling algorithm for online multi-person tracking. Instead of only considering the spatial structure of dynamic appearances in the feature space, we use the HMM to learn a temporal dynamic appearance model (TDAM) to exploit the temporal dependencies between successive appearances. The core of the TDAM lies on the online learning of model parameters as the appearance sequences during tracking are built sequentially. To accomplish online learning, we has proposed a feature selection approach to find an appropriate mid-level semantic feature space, and presented a principled incremental learning algorithm which alternatively evaluates new appearances and adjusts model parameters. We have demonstrated that the TDAM is able to capture both the spatial structure and the temporal dynamic of varying appearances, and thus produces reliable appearance affinities between trajectories and detections.

As a result, the TDAM, in terms of multi-person tracking, does better at preserving identity than previous methods when frequent and close interactions between persons occur. Furthermore, by using effective feature selection and incremental learning approach, the TDAM can be embedded into various online tracking frameworks to improve performance, such as joint probabilistic data association filters and particle filters. Our work is one of the very few attempts that aims to improve multi-person tracking performance by modeling temporal dynamic properties of appearance changes. We hope that our results are inspiring for other researcher to discover this direction more. In future work, we will focus on discriminative learning of the TDAM to capture the correlation between person-specific models. While the TDAM is learned independently for each person in this paper, the dependencies across different models is helpful to further facilitate data association.

%Using the TDAM, we have shown improved multi-person tracking results on various real-world challenging videos. Our experiments also have indicated that our method outperforms the state-of-the-art online trackers and compares favorably against most offline algorithms.

%\section*{Acknowledgements}
%
%This work was supported in part by the Natural Science Foundation of China (NSFC) under Grant 61375044, and in part by the Specialized Research Fund for the Doctoral Program of Higher Education of China under Grant 20121101110035.

{\small
\bibliographystyle{ieee}
\bibliography{egbib}

\begin{thebibliography}{10}\itemsep=-1pt

\bibitem{BaeY2014robust}
S.~Bae and K.~Yoon.
\newblock Robust online multi-object tracking based on tracklet confidence and
  online discriminative appearance learning.
\newblock In {\em CVPR}, pages 1218--1225, 2014.

\bibitem{bae2014robust}
S.~Bae and K.~Yoon.
\newblock Robust online multi-object tracking with data association and track
  management.
\newblock {\em TIP}, 23(7):2820--2833, 2014.

\bibitem{ben2011tracking}
H.~Ben~Shitrit, J.~Berclaz, F.~Fleuret, and P.~Fua.
\newblock Tracking multiple people under global appearance constraints.
\newblock In {\em ICCV}, pages 137--144, 2011.

\bibitem{bhattacharya2014recognition}
S.~Bhattacharya, M.~M. Kalayeh, R.~Sukthankar, and M.~Shah.
\newblock Recognition of complex events: Exploiting temporal dynamics between
  underlying concepts.
\newblock In {\em CVPR}, pages 2243--2250, 2014.

\bibitem{breitenstein2011online}
M.~D. Breitenstein, F.~Reichlin, B.~Leibe, E.~Koller-Meier, and L.~Van~Gool.
\newblock Online multiperson tracking-by-detection from a single, uncalibrated
  camera.
\newblock {\em TPAMI}, 33(9):1820--1833, 2011.

\bibitem{brendel2011multiobject}
W.~Brendel, M.~Amer, and S.~Todorovic.
\newblock Multiobject tracking as maximum weight independent set.
\newblock In {\em CVPR}, pages 1273--1280, 2011.

\bibitem{choi2015near}
W.~Choi.
\newblock Near-online multi-target tracking with aggregated local flow
  descriptor.
\newblock {\em arXiv preprint arXiv:1504.02340}, 2015.

\bibitem{dalal2005histograms}
N.~Dalal and B.~Triggs.
\newblock Histograms of oriented gradients for human detection.
\newblock In {\em CVPR}, pages 886--893, 2005.

\bibitem{dicle2013way}
C.~Dicle, O.~I. Camps, and M.~Sznaier.
\newblock The way they move: Tracking multiple targets with similar appearance.
\newblock In {\em ICCV}, pages 2304--2311, 2013.

\bibitem{digalakis1999online}
V.~V. Digalakis.
\newblock Online adaptation of hidden markov models using incremental
  estimation algorithms.
\newblock {\em IEEE Transactions on Speech and Audio Processing},
  7(3):253--261, 1999.

\bibitem{dollar2014fast}
P.~Doll{\'a}r, R.~Appel, S.~Belongie, and P.~Perona.
\newblock Fast feature pyramids for object detection.
\newblock {\em TPAMI}, 36(8):1532--1545, 2014.

\bibitem{felzenszwalb2010object}
P.~F. Felzenszwalb, R.~B. Girshick, D.~McAllester, and D.~Ramanan.
\newblock Object detection with discriminatively trained part-based models.
\newblock {\em TPAMI}, 32(9):1627--1645, 2010.

\bibitem{geiger20143d}
A.~Geiger, M.~Lauer, C.~Wojek, C.~Stiller, and R.~Urtasun.
\newblock 3{D} traffic scene understanding from movable platforms.
\newblock {\em TPAMI}, 36(5):1012--1025, 2014.

\bibitem{hariharan2012discriminative}
B.~Hariharan, J.~Malik, and D.~Ramanan.
\newblock Discriminative decorrelation for clustering and classification.
\newblock In {\em ECCV}, pages 459--472. 2012.

\bibitem{juang1986maximum}
B.-H. Juang, S.~E. Levinson, and M.~M. Sondhi.
\newblock Maximum likelihood estimation for multivariate mixture observations
  of markov chains.
\newblock {\em IEEE Transactions on Information Theory}, 32(2):307--309, 1986.

\bibitem{keni2008evaluating}
B.~Keni and S.~Rainer.
\newblock Evaluating multiple object tracking performance: the {CLEAR MOT}
  metrics.
\newblock {\em EURASIP Journal on Image and Video Processing}, 2008:1--10,
  2008.

\bibitem{kim2013online}
S.~Kim, S.~Kwak, J.~Feyereisl, and B.~Han.
\newblock Online multi-target tracking by large margin structured learning.
\newblock In {\em ACCV}, pages 98--111. 2013.

\bibitem{kuhn1955hungarian}
H.~W. Kuhn.
\newblock The hungarian method for the assignment problem.
\newblock {\em Naval Research Logistics Quarterly}, 2(1-2):83--97, 1955.

\bibitem{kuo2010multi}
C.-H. Kuo, C.~Huang, and R.~Nevatia.
\newblock Multi-target tracking by on-line learned discriminative appearance
  models.
\newblock In {\em CVPR}, pages 685--692, 2010.

\bibitem{kuo2011does}
C.-H. Kuo and R.~Nevatia.
\newblock How does person identity recognition help multi-person tracking?
\newblock In {\em CVPR}, pages 1217--1224, 2011.

\bibitem{leal2014learning}
L.~Leal-Taix{\'e}, M.~Fenzi, A.~Kuznetsova, B.~Rosenhahn, and S.~Savarese.
\newblock Learning an image-based motion context for multiple people tracking.
\newblock In {\em CVPR}, pages 3542--3549, 2014.

\bibitem{leal2015motchallenge}
L.~Leal-Taix{\'e}, A.~Milan, I.~Reid, S.~Roth, and K.~Schindler.
\newblock {MOTC}hallenge 2015: {T}owards a benchmark for multi-target tracking.
\newblock {\em arXiv preprint arXiv:1504.01942}, 2015.

\bibitem{lee2003video}
K.-C. Lee, J.~Ho, M.-H. Yang, and D.~Kriegman.
\newblock Video-based face recognition using probabilistic appearance
  manifolds.
\newblock In {\em CVPR}, pages 306--313, 2003.

\bibitem{li2013recognizing}
W.~Li, Q.~Yu, H.~Sawhney, and N.~Vasconcelos.
\newblock Recognizing activities via bag of words for attribute dynamics.
\newblock In {\em CVPR}, pages 2587--2594, 2013.

\bibitem{li2009learning}
Y.~Li, C.~Huang, and R.~Nevatia.
\newblock Learning to associate: Hybridboosted multi-target tracker for crowded
  scene.
\newblock In {\em CVPR}, pages 2953--2960, 2009.

\bibitem{lim2005caratheodory}
H.~Lim, O.~I. Camps, and M.~Sznaier.
\newblock A caratheodory-fejer approach to dynamic appearance modeling.
\newblock In {\em CVPR}, pages 301--307, 2005.

\bibitem{lim2006dynamic}
H.~Lim, O.~I. Camps, M.~Sznaier, and V.~I. Morariu.
\newblock Dynamic appearance modeling for human tracking.
\newblock In {\em CVPR}, pages 751--757, 2006.

\bibitem{liu2003video}
X.~Liu and T.~Chen.
\newblock Video-based face recognition using adaptive hidden markov models.
\newblock In {\em CVPR}, pages 333--340, 2003.

\bibitem{milan2015joint}
A.~Milan, L.~Leal-Taix{\'e}, K.~Schindler, and I.~Reid.
\newblock Joint tracking and segmentation of multiple targets.
\newblock In {\em CVPR}, 2015.

\bibitem{Milan:2014:CEM}
A.~Milan, S.~Roth, and K.~Schindler.
\newblock Continuous energy minimization for multitarget tracking.
\newblock {\em TPAMI}, 36(1):58--72, 2014.

\bibitem{pirsiavash2011globally}
H.~Pirsiavash, D.~Ramanan, and C.~C. Fowlkes.
\newblock Globally-optimal greedy algorithms for tracking a variable number of
  objects.
\newblock In {\em CVPR}, pages 1201--1208, 2011.

\bibitem{poiesi2013multi}
F.~Poiesi, R.~Mazzon, and A.~Cavallaro.
\newblock Multi-target tracking on confidence maps: {A}n application to people
  tracking.
\newblock {\em Computer Vision and Image Understanding}, 117(10):1257--1272,
  2013.

\bibitem{rabiner1989tutorial}
L.~Rabiner.
\newblock A tutorial on hidden markov models and selected applications in
  speech recognition.
\newblock {\em Proceedings of the IEEE}, 77(2):257--286, 1989.

\bibitem{roweis2000nonlinear}
S.~T. Roweis and L.~K. Saul.
\newblock Nonlinear dimensionality reduction by locally linear embedding.
\newblock {\em Science}, 290(5500):2323--2326, 2000.

\bibitem{shi2000normalized}
J.~Shi and J.~Malik.
\newblock Normalized cuts and image segmentation.
\newblock {\em TPAMI}, 22(8):888--905, 2000.

\bibitem{shu2012part}
G.~Shu, A.~Dehghan, O.~Oreifej, E.~Hand, and M.~Shah.
\newblock Part-based multiple-person tracking with partial occlusion handling.
\newblock In {\em CVPR}, pages 1815--1821, 2012.

\bibitem{wang2014tracklet}
B.~Wang, G.~Wang, K.~L. Chan, and L.~Wang.
\newblock Tracklet association with online target-specific metric learning.
\newblock In {\em CVPR}, pages 1234--1241, 2014.

\bibitem{wu2007detection}
B.~Wu and R.~Nevatia.
\newblock Detection and tracking of multiple, partially occluded humans by
  bayesian combination of edgelet based part detectors.
\newblock {\em IJCV}, 75(2):247--266, 2007.

\bibitem{xing2009multi}
J.~Xing, H.~Ai, and S.~Lao.
\newblock Multi-object tracking through occlusions by local tracklets filtering
  and global tracklets association with detection responses.
\newblock In {\em CVPR}, pages 1200--1207, 2009.

\bibitem{yang2012multi}
B.~Yang and R.~Nevatia.
\newblock Multi-target tracking by online learning of non-linear motion
  patterns and robust appearance models.
\newblock In {\em CVPR}, pages 1918--1925, 2012.

\bibitem{yang2009detection}
M.~Yang, F.~Lv, W.~Xu, and Y.~Gong.
\newblock Detection driven adaptive multi-cue integration for multiple human
  tracking.
\newblock In {\em ICCV}, pages 1554--1561, 2009.

\bibitem{yoon2015bayesian}
J.~H. Yoon, M.-H. Yang, J.~Lim, and K.-J. Yoon.
\newblock Bayesian multi-object tracking using motion context from multiple
  objects.
\newblock In {\em WACV}, pages 33--40, 2015.

\end{thebibliography}
}

\end{document}